\pdfoutput=1
\documentclass[article]{IEEEtran}
\usepackage{graphicx}
\usepackage[table,xcdraw]{xcolor}
\usepackage{adjustbox}
\usepackage{tabularx, booktabs}
\usepackage{tablefootnote}
\usepackage{threeparttable}

\usepackage[space]{cite}

\usepackage{tikz}
\usepackage[cmex10]{amsmath}
\usepackage{amsfonts}
\usepackage{algorithm}
\usepackage[noend]{algpseudocode}
\usepackage{soul}
\usepackage[framemethod=tikz]{mdframed}

\usepackage{array}

\usepackage{mdwmath}

\usepackage{eqparbox}

\usepackage[]{caption}
\usepackage{hyperref}
\hypersetup{
    colorlinks=false,
    linkcolor=black,
    filecolor=magenta,      
    urlcolor=black,
    citecolor=blue,
}
\soulregister\cite7
\soulregister\ref7
\soulregister\pageref7

\colorlet{usercolorname}{white}
\sethlcolor{usercolorname}

\begin{document}
%
\title{Insights into LSTM Fully Convolutional Networks for Time Series Classification}
%
%
\makeatletter
\let\@fnsymbol\@arabic
\makeatother

\author{{Fazle~Karim$^{1 *}$,~\IEEEmembership{Graduate Student Member,~IEEE},
        Somshubra~Majumdar$^{2 *}$
        \\Houshang~Darabi$^{1}$,~\IEEEmembership{Senior Member,~IEEE}
        }
\thanks{$^{*}$Equal contribution.}
\thanks{$^{1}$Mechanical and Industrial Engineering, University of Illinois at Chicago, Chicago, IL}
\thanks{$^{2}$Computer Science, University of Illinois at Chicago, Chicago, IL}}
\maketitle

\begin{abstract}
Long Short Term Memory Fully Convolutional Neural Networks (LSTM-FCN) and Attention LSTM-FCN (ALSTM-FCN) have shown to achieve state-of-the-art performance on the task of classifying time series signals on the old University of California-Riverside (UCR) time series repository. However, there has been no study on why LSTM-FCN and ALSTM-FCN perform well. In this paper, we perform a series of ablation tests (3627 experiments) on LSTM-FCN and ALSTM-FCN to provide a better understanding of the model and each of its sub-module. Results from the ablation tests on ALSTM-FCN and LSTM-FCN show that the LSTM and the FCN blocks perform better when applied in a conjoined manner. Two z-normalizing techniques, z-normalizing each sample independently and z-normalizing the whole dataset, are compared using a Wilcoxson signed-rank test to show a statistical difference in performance. In addition, we provide an understanding of the impact dimension shuffle has on LSTM-FCN by comparing its performance with LSTM-FCN when no dimension shuffle is applied. Finally, we demonstrate the performance of the LSTM-FCN when the LSTM block is replaced by a GRU, basic RNN, and Dense Block. 
\end{abstract}


%
\IEEEpeerreviewmaketitle

\section{Introduction}

\bstctlcite{IEEEexample:BSTcontrol}

Time series classification has recently received a lot of attention over the past three decades \cite{dau2018ucr,sharabiani2017efficient,Sharabiani2014_bayesian,karim2018multivariate}. Such data is widely available everywhere and collected with various sensors \cite{wei2006semi}. A variety of real world sensors capture time series information such as weather readings \cite{taylor2009wind}, stock market data\cite{tsay2005analysis}, and EEG / ECG \cite{sternickel2002automatic,theiler1992testing}. \hl{Time series classification is a supervised learning task that classifies a series of data points that are commonly collected in equal intervals and depicted in a sequential order \cite{maharaj2019time}. Typically, the input to a time series classification problem is a time series signal, $X \in \mathbb{R}^{T \times F}$, such that $X_{t} \in \mathbb{R}^{F}$ is the input feature vector of length $F$ at time step $t$, where $0 < t \leq T$. The maximum length of each time series, $T$, may vary \cite{Lea_2016}. The output of a time series classification problem, $Y \in \{1, \dots, C\}$, is a discrete class/category label that represents the input time series signals. The total number of classes, $C$, is dependent on the time series classification problem. The main challenges faced in time series classification are how to efficiently (speed and space) \cite{xi2006fast} and effectively (accurately) \cite{jain2018asymmetric} classify a time series. }

Some of the earliest work that applies data mining techniques for time series classification dates back to the early 1990s when authors would apply various algorithms onto single artificial datasets \cite{agrawal1993efficient,keogh2003need}. Since the initial decade of research in this field, \textit{Chen et al.}\cite{UCRArchive} have graciously helped the community by collecting and making 85 time series datasets from various domains available online to the public for research purposes. This has lead to rapid progress in the field of time series classification and yielded a significant body of work. Recently, \textit{Dau et al.} \cite{UCRArchive2018} have updated the repository with 43 datasets with time series datasets. These datasets have a significantly higher number of samples, several of which have long time dependencies or incorporates variable sequence lengths, which makes the task of sequence classification far more exigent. Most of the new datasets also have a significantly larger test set and a few have variable time series lengths to represent real-world scenarios \cite{dau2018ucr}.

Several researchers have used the old archive benchmark datasets to propose feature-based models \cite{Lin_2007,Baydogan_2013,Sch_fer_2014,schafer2016scalable,Schafer_2017}, ensembles \cite{Lines_2014,bagnall2015time} and deep learning models \cite{cui2016multi,wang2017time,karim2018lstm} to \hl{accurately} classify the time series data. One of the current state-of-the-art models that classify the time series datasets from the repository developed by \textit{Chen et al.} \cite{UCRArchive} are the Long Short Term Memory Fully Convolutional Network (LSTM-FCN) and the Attention LSTM-FCN proposed by \textit{Karim \& Majumdar et al.} \cite{karim2018lstm}. LSTM-FCN and ALSTM-FCN are deep learning models, a Fully Convolutional Network (FCN) module augmented with a Long Term Short Term Recurrent Neural Network (LSTM) that classify time series datasets. \hl{LSTM-FCN and ALSTM-FCN have received a lot of attention from the time series classification community due to their advantage over other models. In terms of classification accuracy, both the models outperform several traditional time series classification models, while requiring minimal pre-processing of the data. A significant advantage of utilizing these models is their ability to compute features on their own, eliminating the requirement for significant domain expertise and manual feature extraction. Furthermore, both these models can easily scale with a larger amount of time series data, which is generated daily by automated processes. Finally, LSTM-FCN has already been deployed in real world scenarios. One such application is to efficiently classify pet dog sounds using resource constrained sensors \cite{kim2018resource}.} The original models, LSTM-FCN and ALSTM-FCN, lacked the explanation of each sub-module. In this paper, we provide detailed ablation tests to explain the sub-modules of the models. 

The remainder of the paper is organized as follows. Section \ref{Experiments} presents the parameters used in developing the models and discusses the experiments performed. Section \ref{Results} compares two \textit{z-normalization} schemes.  Subsequently, Section \ref{AblationTests} provides a detailed ablation test on the deep learning models, LSTM-FCN and ALSTM. Finally, Section \ref{conclusion} concludes the paper.

 \vspace{-1mm}
\begin{mdframed}[hidealllines=true,backgroundcolor=white]

\section{Background Review}
\label{BackgroundReview}

\subsection{Temporal Convolutions Networks}
Temporal convolution network is a type of artificial neural network whose input is generally a time series signal, $X$, where $X_{t} \in \mathbb{R}^{F}$ is the input feature vector of length $F$ for time step $t$ for $0 < t \leq T$. $T$ may vary for each time series sequence \cite{Lea_2016}.

In a temporal convolution network, 1D filters are applied on each convolutional layer, $L$, that discovers the evolution of the input signal over the course of an action. \textit{Lea et al.} \cite{Lea_2016} discusses each filter of each layer are parameterized by tensor $W^{(l)} \in \mathbb{R}^{F_l \times d \times F_{l-1}} $ and biases $b^{(l)} \in \mathbb{R}^{F_l}$, where $l \in \{1, . . . , L\}$ is the layer index and $d$ is the filter duration. The $i$-th element of the activation ${\mathbf {\hat{E}}}^{(l)}_{t} \in \mathbb{R}^{F_{l}}$ of the $l$-th layer is a function of the activation matrix $E^{(l-1)} \in \mathbb{R}^{F_{l-1} \times T_{l-1}}$ of the previous layer, such that, 

\begin{equation}
{\mathbf {\hat{E}}}_{i, t}^{(l)} = f\left(b_{i}^{(l)} + \sum_{t'=1}^{d} \left< W_{i, t', .}^{(l)}, E_{., t+d-t'}^{(l-1)} \right> \right)
\end{equation}
for each time $t$ where $f(\cdot)$ is a Rectified Linear Unit.

Typically, a convolutional layer is followed by batch normalization \cite{ioffe2015batch}. Subsequently, this is trailed by an activation function (a Rectified Linear Unit or a Parametric Rectified Linear Unit \cite{Trottier2016}).

\subsection{Recurrent Neural Networks}
\def\x{{\mathbf x}}
\def\L{{\cal L}}

Recurrent Neural Networks (RNN) are a type of artificial neural network that demonstrates stateful temporal behavior given a time sequence. \textit{Pascanu et al.} \cite{pascanu2013construct} proposed an RNN to preserve a hidden vector $\mathbf h$ as a state that is updated at time step $t$,
\begin{equation}
	\mathbf h_t = \tanh(\mathbf W\mathbf h_{t-1} + \mathbf I\mathbf  \x_t),
\end{equation}
where $tanh$ is the hyperbolic tangent function, $\mathbf x_t$ is the input vector at time step $t$, $\mathbf W$ is the recurrent weight matrix and $\mathbf I$ is the projection matrix. The prediction, $\mathbf y_t$, is computed such that, 
\begin{equation}
	\mathbf y_t = \text{softmax}(\mathbf W\mathbf h_{t-1}),
\end{equation}
where $\mathbf h$ is a hidden state, $\mathbf W$ is a weight matrix and softmax operation  normalizes the output of the model to a valid probability distribution and the logistic sigmoid function is shown as $\sigma$. Deep RNNs can be formed by stacking the output of one RNN as the input to another, such that the hidden state,  $\mathbf h^{l-1}$ of a RNN layer $l{-1}$, is an input to the hidden state, $\mathbf h^l$ of another RNN layer $l$. In other words,  
\begin{equation}
	\mathbf h_t^{l} = \sigma(\mathbf W\mathbf h_{t-1}^{l} + \mathbf I\mathbf h_t^{l-1}).
\end{equation}

RNNs are prone to be affected by vanishing gradients. This issue is addressed using a Long short-term memory (LSTM) or a Gated Recurrent Unit (GRU).

\subsection{Long Short-Term Memory RNNs}
\def\x{{\mathbf x}}
To solve the vanishing gradient problem, LSTM RNNs utilize gating functions in their state dynamics \cite{hochreiter1997long}. Each LSTM cell contains a hidden vector, $\mathbf h$, and a memory vector, $\mathbf m$. At each time step, the memory vector regulates the state updates and outputs, such that the following computation is performed computed as follows (first depicted by \textit{Graves et al.} \cite{graves2012supervised}): 

\begin{equation}
	\begin{split}
		& \mathbf g^u = \sigma(\mathbf W^u\mathbf h_{t-1}  + \mathbf I^u\x_t ) \\
		& \mathbf g^f = \sigma(\mathbf W^f\mathbf h_{t-1} + \mathbf I^f\x_t) \\
		& \mathbf g^o = \sigma(\mathbf W^o\mathbf h_{t-1} + \mathbf I^o\x_t) \\
		& \mathbf g^c = \tanh(\mathbf W^c\mathbf h_{t-1} + \mathbf I^c\x_t) \\
		& \mathbf m_t = \mathbf g^f \odot \mathbf m_{t-1}\mathbf \ + \  \mathbf g^u \odot 
		\mathbf g^c \\
		& \mathbf h_t = \tanh(\mathbf g^o \odot \mathbf m_t) 
	\end{split}
\end{equation}
where \textbf{$\mathbf g^u$}, \textbf{$\mathbf g^f$}, \textbf{$\mathbf g^o$}, \textbf{$\mathbf g^c$} are the activation vectors of the input, forget, output and cell state gates respectively, $\mathbf W^u, \mathbf W^f, \mathbf W^o, \mathbf W^c$ are the recurrent weight matrices, $\mathbf I^u, \mathbf I^f, \mathbf I^o, \mathbf I^c$ portrays the projection matrices, $\sigma$ is the logistic sigmoid function, $\odot$ is an elementwise multiplication, and $h_t$ is the hidden state vector of the $t$th time step.

\subsection{Gated Recurrent Unit}
Cho et al. \cite{cho2014gru} proposed a modification to the LSTM RNN that also solves the vanishing gradient problem using an update and reset gate. Due to the simpler gating structure of the model, reduced number of gates and thereby parameters, it is considered to be an efficient alternative to the LSTM RNN.

\begin{align}
    \mathbf{z_t} &= \sigma_g{(\mathbf{W_z x_t} + \mathbf{U_z h_{t-1}})} \\
    \mathbf{r_t} &= \sigma_g{(\mathbf{W_r x_t} + \mathbf{U_r h_{t-1}})} \\
    \mathbf{h_t} &= (1 - \mathbf{z_t}) \odot \mathbf{h_{t-1}} \: + \\ \nonumber 
    &\mathbf{z_t} \odot \sigma_h{(\mathbf{W_h x_t} + \mathbf{U_h (r_t \odot h_{t-1})})}
\end{align}

where $x_t$ is the input vector at time step $t$, $z_t$ is the update gate vector, $r_t$ is the reset gate vector, $h_t$ is the hidden state and output vector, $W_z$ and $W_r$ are the trainable weight matrices for the update and reset gate respectively, $U_z$ and $U_r$ are the trainable recurrent weight matrices for the update and reset gate respectively, $\sigma_g$ is the logistic sigmoid function,  $\sigma_h$ is the hyperbolic tangent function and $\odot$ is the Hadamard product of the two inputs.

\subsection{Fully Connected (Dense) Layer}
A fully connected layer can be described as a dense matrix multiplication of the input vector with a trainable weight matrix, and optionally, the addition of a trainable bias vector to the output. The output of each layer can be represented by:

\begin{equation}
    output = a \,(\mathbf{W x} + b)
\end{equation}

where $\mathbf{W}$ is a weight matrix, $b$ is a bias vector, and $a$ is a non-linear activation function. Common activation functions are the Rectified Linear Unit (ReLU), the logistic sigmoid function, or a hyperbolic tangent function.

\end{mdframed}
\section{Experiments}
\label{Experiments}
The LSTM-FCN and ALSTM-FCN models are trained on various released UCR benchmark datasets. The benchmark datasets include a train and test set which is used for model training and validation. We utilize the same structure of the models as the original models \cite{karim2018lstm} and perform grid search to find the optimal number of LSTM cells from the set consisting of 8, 64 or 128 cells. All models are trained for 2000 epochs. The batch size of 128 is kept consistent for all datasets. All LSTM or Attention LSTM layers are followed by dropout layer with a probability of 80 percent to prevent overfitting. Class imbalance is handled via a class weighing scheme inspired by King and Zeng \cite{king2001logistic}. All models are trained using the Keras library \cite{chollet2015keras} with Tensorflow \cite{tensorflow2015-whitepaper} as the backend and are made available publically\footnote{The codes and weights of each model are made available at \href{https://github.com/houshd/LSTM-FCN}{https://github.com/houshd/LSTM-FCN-Ablation}}.

All models are trained via gradient descent using the Adam optimizer \cite{kingma2014adam}. The initial learning rate was set to 1\textit{e-3} and is reduced to a minimum of 1\textit{e-4}. We reduced the learning rate by a factor of $1/{\sqrt[3]{2}}$, whenever the training loss of 100 consecutive epochs do not improve. The model weights are updated only through the training loss. The accuracies we report are based on the best models we find. The methodology we follow is common in various deep learning applications \cite{Krizhevsky2012Imagenet,he2015delving,Szegedy_2016_CVPR,szegedy2017inception,hu2018senet}. In addition, we utilize the initialization proposed by \textit{He et al.} \cite{he2015delving} for all convolutional layers. The input data is \textit{z-normalized} and the datasets with variable length time series are padded with zeros at the end to match the longest time series in that dataset. All models are evaluated using classification accuracy and \textit{mean-per-class-error} (MPCE), which is defined as the average error of each class for all the datasets and mathematically represented as following:  

\begin{align*}
 \label{eq1}
PCE_k &=\frac{1-accuracy}{\textit{number of unique classes}} \\
MPCE &=\frac{1}{\textit{N}} \sum_{k=1}^N {PCE_k}.    
\end{align*}

\hspace{0.08cm}


\section{Dataset Ablation Test}

 \newcolumntype{C}{>{\centering\arraybackslash}X}
 \newcommand\mcxl[1]{\multicolumn{1}{|C|}{\bfseries #1}}
 \newcommand\mcx[1]{\multicolumn{1}{C }{\bfseries #1}}
 


 \begin{table}[]
 \centering
 \caption{Performance comparison of LSTM-FCN and ALSTM-FCN with the baseline models. Green cells designate instances where our performance matches or exceeds state-of-the-art results. Bold values denote model with the best performance.}

\label{tab:perf_tab}
\begin{adjustbox}{width=1 \linewidth}

 \begin{tabularx}{0.62 \textwidth}{|C|C|C|C|C|C|}
    \hline
    Name & Baseline \cite{dau2018ucr} & LSTM-FCN Data Normalized & ALSTM-FCN Data Normalized & LSTM-FCN Sample Normalized & ALSTM-FCN Sample Normalized \\
    \hline
    ACSF1 & 0.6400 & \cellcolor[rgb]{ .663,  .816,  .557}\textbf{0.9300} & \cellcolor[rgb]{ .663,  .816,  .557}0.9100 & \cellcolor[rgb]{ .663,  .816,  .557}0.9200 & \cellcolor[rgb]{ .663,  .816,  .557}0.9200 \\
    \hline
    AllGestWiX & 0.7171 & \cellcolor[rgb]{ .663,  .816,  .557}\textbf{0.7214} & \cellcolor[rgb]{ .663,  .816,  .557}0.7200 & 0.7071 & 0.7086 \\
    \hline
    AllGestWiY & 0.7300 & \cellcolor[rgb]{ .663,  .816,  .557}0.7786 & \cellcolor[rgb]{ .663,  .816,  .557}0.7914 & \cellcolor[rgb]{ .663,  .816,  .557}\textbf{0.7929} & \cellcolor[rgb]{ .663,  .816,  .557}0.7829 \\
    \hline
    AllGestWiZ & 0.6514 & \cellcolor[rgb]{ .663,  .816,  .557}\textbf{0.7400} & \cellcolor[rgb]{ .663,  .816,  .557}0.7357 & \cellcolor[rgb]{ .663,  .816,  .557}0.6800 & \cellcolor[rgb]{ .663,  .816,  .557}0.6914 \\
    \hline
    BME  & 0.9800 & \cellcolor[rgb]{ .663,  .816,  .557}\textbf{1.0000} & 0.8333 & \cellcolor[rgb]{ .663,  .816,  .557}0.9933 & 0.8600 \\
    \hline
    Chinatown & 0.9565 & \cellcolor[rgb]{ .663,  .816,  .557}\textbf{0.9855} & \cellcolor[rgb]{ .663,  .816,  .557}\textbf{0.9855} & \cellcolor[rgb]{ .663,  .816,  .557}0.9826 & \cellcolor[rgb]{ .663,  .816,  .557}0.9797 \\
    \hline
    Crop & 0.7117 & \cellcolor[rgb]{ .663,  .816,  .557}\textbf{0.7652} & \cellcolor[rgb]{ .663,  .816,  .557}0.7638 & \cellcolor[rgb]{ .663,  .816,  .557}0.7425 & \cellcolor[rgb]{ .663,  .816,  .557}0.7389 \\
    \hline
    DodgLpDay & 0.5875 & \cellcolor[rgb]{ .663,  .816,  .557}\textbf{0.6375} & 0.4875 & \cellcolor[rgb]{ .663,  .816,  .557}0.6125 & \cellcolor[rgb]{ .663,  .816,  .557}0.5875 \\
    \hline
    DodgLpGm & \cellcolor[rgb]{ .663,  .816,  .557}\textbf{0.9275} & 0.8913 & 0.7754 & 0.8986 & 0.8261 \\
    \hline
    DodgLpWnd & 0.9855 & \cellcolor[rgb]{ .663,  .816,  .557}\textbf{0.9855} & 0.9710 & 0.9783 & 0.9275 \\
    \hline
    EOGHzSgn & 0.5028 & \cellcolor[rgb]{ .663,  .816,  .557}0.6547 & \cellcolor[rgb]{ .663,  .816,  .557}\textbf{0.6878} & \cellcolor[rgb]{ .663,  .816,  .557}0.6409 & \cellcolor[rgb]{ .663,  .816,  .557}0.6133 \\
    \hline
    EOGVtSgn & 0.4751 & \cellcolor[rgb]{ .663,  .816,  .557}\textbf{0.5387} & \cellcolor[rgb]{ .663,  .816,  .557}0.5138 & \cellcolor[rgb]{ .663,  .816,  .557}0.5028 & 0.4696 \\
    \hline
    EthLevel & 0.2820 & \cellcolor[rgb]{ .663,  .816,  .557}\textbf{0.7660} & \cellcolor[rgb]{ .663,  .816,  .557}0.7380 & \cellcolor[rgb]{ .663,  .816,  .557}\textbf{0.7660} & \cellcolor[rgb]{ .663,  .816,  .557}0.7480 \\
    \hline
    FrzRegTr & 0.9070 & \cellcolor[rgb]{ .663,  .816,  .557}0.9986 & \cellcolor[rgb]{ .663,  .816,  .557}\textbf{0.9989} & \cellcolor[rgb]{ .663,  .816,  .557}\textbf{0.9989} & \cellcolor[rgb]{ .663,  .816,  .557}\textbf{0.9989} \\
    \hline
    FrzSmlTr & 0.7533 & \cellcolor[rgb]{ .663,  .816,  .557}0.8295 & \cellcolor[rgb]{ .663,  .816,  .557}\textbf{0.8747} & \cellcolor[rgb]{ .663,  .816,  .557}0.8025 & \cellcolor[rgb]{ .663,  .816,  .557}0.8407 \\
    \hline
    Fungi & 0.8387 & \cellcolor[rgb]{ .663,  .816,  .557}\textbf{1.0000} & \cellcolor[rgb]{ .663,  .816,  .557}0.9946 & \cellcolor[rgb]{ .663,  .816,  .557}0.9892 & \cellcolor[rgb]{ .663,  .816,  .557}0.9839 \\
    \hline
    GestMidAirD1 & 0.6385 & \cellcolor[rgb]{ .663,  .816,  .557}\textbf{0.7462} & \cellcolor[rgb]{ .663,  .816,  .557}0.7154 & \cellcolor[rgb]{ .663,  .816,  .557}0.7308 & \cellcolor[rgb]{ .663,  .816,  .557}0.7231 \\
    \hline
    GestMidAirD2 & 0.6077 & \cellcolor[rgb]{ .663,  .816,  .557}0.6923 & \cellcolor[rgb]{ .663,  .816,  .557}0.6385 & \cellcolor[rgb]{ .663,  .816,  .557}\textbf{0.7077} & \cellcolor[rgb]{ .663,  .816,  .557}0.6923 \\
    \hline
    GestMidAirD3 & 0.3769 & \cellcolor[rgb]{ .663,  .816,  .557}\textbf{0.4538} & \cellcolor[rgb]{ .663,  .816,  .557}0.4462 & \cellcolor[rgb]{ .663,  .816,  .557}\textbf{0.4538} & \cellcolor[rgb]{ .663,  .816,  .557}0.3846 \\
    \hline
    GestPebZ1 & 0.8256 & \cellcolor[rgb]{ .663,  .816,  .557}\textbf{0.9419} & \cellcolor[rgb]{ .663,  .816,  .557}0.9244 & \cellcolor[rgb]{ .663,  .816,  .557}\textbf{0.9419} & \cellcolor[rgb]{ .663,  .816,  .557}0.9128 \\
    \hline
    GestPebZ2 & 0.7785 & \cellcolor[rgb]{ .663,  .816,  .557}\textbf{0.8987} & \cellcolor[rgb]{ .663,  .816,  .557}0.8354 & \cellcolor[rgb]{ .663,  .816,  .557}0.8544 & \cellcolor[rgb]{ .663,  .816,  .557}0.8861 \\
    \hline
    GunPtAgeSp & 0.9652 & \cellcolor[rgb]{ .663,  .816,  .557}\textbf{1.0000} & \cellcolor[rgb]{ .663,  .816,  .557}\textbf{1.0000} & \cellcolor[rgb]{ .663,  .816,  .557}0.9968 & \cellcolor[rgb]{ .663,  .816,  .557}\textbf{1.0000} \\
    \hline
    GunPointMVsF & 0.9968 & \cellcolor[rgb]{ .663,  .816,  .557}\textbf{1.0000} & \cellcolor[rgb]{ .663,  .816,  .557}\textbf{1.0000} & \cellcolor[rgb]{ .663,  .816,  .557}\textbf{1.0000} & \cellcolor[rgb]{ .663,  .816,  .557}\textbf{1.0000} \\
    \hline
    GunPointOVsY & 0.9651 & \cellcolor[rgb]{ .663,  .816,  .557}\textbf{1.0000} & \cellcolor[rgb]{ .663,  .816,  .557}\textbf{1.0000} & \cellcolor[rgb]{ .663,  .816,  .557}0.9968 & \cellcolor[rgb]{ .663,  .816,  .557}0.9968 \\
    \hline
    HouseTwenty & 0.9412 & \cellcolor[rgb]{ .663,  .816,  .557}0.9664 & \cellcolor[rgb]{ .663,  .816,  .557}0.9496 & \cellcolor[rgb]{ .663,  .816,  .557}\textbf{0.9832} & \cellcolor[rgb]{ .663,  .816,  .557}\textbf{0.9832} \\
    \hline
    InsEPGRegTr & 0.8715 & \cellcolor[rgb]{ .663,  .816,  .557}\textbf{1.0000} & \cellcolor[rgb]{ .663,  .816,  .557}\textbf{1.0000} & \cellcolor[rgb]{ .663,  .816,  .557}0.9960 & \cellcolor[rgb]{ .663,  .816,  .557}\textbf{1.0000} \\
    \hline
    InsEPGSmlTr & 0.7349 & \cellcolor[rgb]{ .663,  .816,  .557}\textbf{1.0000} & \cellcolor[rgb]{ .663,  .816,  .557}\textbf{1.0000} & \cellcolor[rgb]{ .663,  .816,  .557}0.9478 & \cellcolor[rgb]{ .663,  .816,  .557}0.9438 \\
    \hline
    MelbPed & 0.8482 & \cellcolor[rgb]{ .663,  .816,  .557}0.9747 & \cellcolor[rgb]{ .663,  .816,  .557}\textbf{0.9755} & \cellcolor[rgb]{ .663,  .816,  .557}0.9147 & \cellcolor[rgb]{ .663,  .816,  .557}0.9135 \\
    \hline
    MxShpRegTr & 0.9089 & \cellcolor[rgb]{ .663,  .816,  .557}\textbf{0.9748} & \cellcolor[rgb]{ .663,  .816,  .557}0.9720 & \cellcolor[rgb]{ .663,  .816,  .557}0.9711 & \cellcolor[rgb]{ .663,  .816,  .557}0.9678 \\
    \hline
    MxShpSmlTr & 0.8355 & \cellcolor[rgb]{ .663,  .816,  .557}0.9365 & \cellcolor[rgb]{ .663,  .816,  .557}0.9274 & \cellcolor[rgb]{ .663,  .816,  .557}\textbf{0.9390} & \cellcolor[rgb]{ .663,  .816,  .557}0.9225 \\
    \hline
    PickGestWiZ & 0.6600 & \cellcolor[rgb]{ .663,  .816,  .557}\textbf{0.9000} & \cellcolor[rgb]{ .663,  .816,  .557}\textbf{0.9000} & \cellcolor[rgb]{ .663,  .816,  .557}0.7600 & \cellcolor[rgb]{ .663,  .816,  .557}0.7000 \\
    \hline
    PigAryPress & 0.1058 & \cellcolor[rgb]{ .663,  .816,  .557}\textbf{0.7885} & \cellcolor[rgb]{ .663,  .816,  .557}0.7596 & \cellcolor[rgb]{ .663,  .816,  .557}0.4231 & \cellcolor[rgb]{ .663,  .816,  .557}0.3942 \\
    \hline
    PigArtPress & 0.2452 & \cellcolor[rgb]{ .663,  .816,  .557}0.9808 & \cellcolor[rgb]{ .663,  .816,  .557}0.9904 & \cellcolor[rgb]{ .663,  .816,  .557}\textbf{1.0000} & \cellcolor[rgb]{ .663,  .816,  .557}\textbf{1.0000} \\
    \hline
    PigCVP & 0.1587 & \cellcolor[rgb]{ .663,  .816,  .557}0.9231 & \cellcolor[rgb]{ .663,  .816,  .557}\textbf{0.9279} & \cellcolor[rgb]{ .663,  .816,  .557}0.8702 & \cellcolor[rgb]{ .663,  .816,  .557}0.8702 \\
    \hline
    PLAID & 0.8399 & 0.4842 & 0.5047 & \cellcolor[rgb]{ .663,  .816,  .557}\textbf{0.9088} & \cellcolor[rgb]{ .663,  .816,  .557}0.8994 \\
    \hline
    PowerCons & 0.9333 & \cellcolor[rgb]{ .663,  .816,  .557}\textbf{1.0000} & \cellcolor[rgb]{ .663,  .816,  .557}0.9500 & \cellcolor[rgb]{ .663,  .816,  .557}0.9944 & 0.9167 \\
    \hline
    Rock & 0.8400 & 0.8000 & 0.8000 & \cellcolor[rgb]{ .663,  .816,  .557}\textbf{0.9200} & \cellcolor[rgb]{ .663,  .816,  .557}0.8600 \\
    \hline
    SgHdGendCh2 & 0.8450 & \cellcolor[rgb]{ .663,  .816,  .557}\textbf{0.9400} & \cellcolor[rgb]{ .663,  .816,  .557}0.8567 & \cellcolor[rgb]{ .663,  .816,  .557}0.9200 & \cellcolor[rgb]{ .663,  .816,  .557}0.8567 \\
    \hline
    SgHdMovCh2 & 0.6378 & \cellcolor[rgb]{ .663,  .816,  .557}\textbf{0.7044} & 0.5622 & 0.5422 & 0.5556 \\
    \hline
    SgHdSubCh2 & 0.8000 & \cellcolor[rgb]{ .663,  .816,  .557}\textbf{0.9222} & 0.6533 & \cellcolor[rgb]{ .663,  .816,  .557}0.8800 & 0.7911 \\
    \hline
    ShkGestWiZ & 0.8600 & \cellcolor[rgb]{ .663,  .816,  .557}\textbf{0.9800} & \cellcolor[rgb]{ .663,  .816,  .557}\textbf{0.9800} & \cellcolor[rgb]{ .663,  .816,  .557}0.9000 & \cellcolor[rgb]{ .663,  .816,  .557}0.8800 \\
    \hline
    SmthSub & 0.9467 & \cellcolor[rgb]{ .663,  .816,  .557}\textbf{1.0000} & \cellcolor[rgb]{ .663,  .816,  .557}\textbf{1.0000} & \cellcolor[rgb]{ .663,  .816,  .557}0.9867 & \cellcolor[rgb]{ .663,  .816,  .557}0.9867 \\
    \hline
    UMD  & 0.9931 & \cellcolor[rgb]{ .663,  .816,  .557}\textbf{1.0000} & 0.9931 & 0.9792 & 0.9722 \\
    \hline
    MPCE & -    & 0.0172 & 0.0242 & 0.0191 & 0.0242 \\
    \hline
    Count & 1 & 30 & 14 & 12 & 6\\
    \hline

    \end{tabularx}%
    \end{adjustbox}
  \label{tab:perform_table}%

\end{table}%


\label{Results}

Table \ref{tab:perform_table} represents the accuracies obtained by applying LSTM-FCN and ALSTM-FCN on the 43 new UCR benchmark datasets based on two \textit{z-normalization} schemes when normalizing the datasets prior to training. These 43 UCR benchmark datasets are the only datasets in the repository that are not padded, normalized or pre-processed in any way. The \textit{dataset mean and standard deviation} is calculated as the mean and standard deviation of only the train set, and then applied to both train and tests, whereas the \textit{sample mean and standard deviation} was calculated for each individual sample separately. When using LSTM-FCN and ALSTM-FCN, our results indicate that when the whole dataset is \textit{z-normalized}, it performs better on 34 datasets (LSTM-FCN) and 30 datasets (ALSTM-FCN) than when each sample is \textit{z-normalized} separately. In addition, a Wilcoxon signed-rank test \cite{wilcoxon1964some} was performed to compare this, yielding a p-value of 4.57\textit{e-07}. We chose the significance level (alpha) of 0.05 for all statistical tests. Since the p-value is less than the \textit{Dunn-Sidak} \cite{abdi2007bonferroni} corrected significance level (alpha) of 0.025, we conclude that \textit{z-normalizing} the whole dataset performs differently than when \textit{z-normalizing} each sample. 

We recommend \textit{z-normalizing} the whole dataset \textit{iff} one knows that the train set can sufficiently represent the global population of the dataset. In other words, if no \textit{a priori} information or domain knowledge is known about the train set, it is safer to \textit{z-normalize} each sample separately, as explained by \textit{Dau et al}. \cite{dau2018ucr}. They provide an example explaining why it is safer to \textit{z-normalize} each sample separately using the dataset \textit{GunPoint}, where a video is converted into a time series. If another video is taken where ``the camera is zoomed in or out, or the actors stood a little closer to the camera, or that the female actor decided to wear new shoes with a high heel'' \cite{dau2018ucr}, the converted time series will be different. The train set will not have this distribution as the validation or test set, and the prediction made by this classifier will be off. In this scenario, it would be best to \textit{z-normalize} each sample separately. On the other hand, if a domain expert knows the train set contains a wide range of samples that represent the different types and amplitudes of time series, \textit{z-normalizing} via the dataset mean and standard deviation would be wiser when using LSTM-FCN and ALSTM-FCN as classifiers.

\section{Model Ablation Tests}

\begin{figure*}[t]
    \centering
    \includegraphics[width=0.75\linewidth]{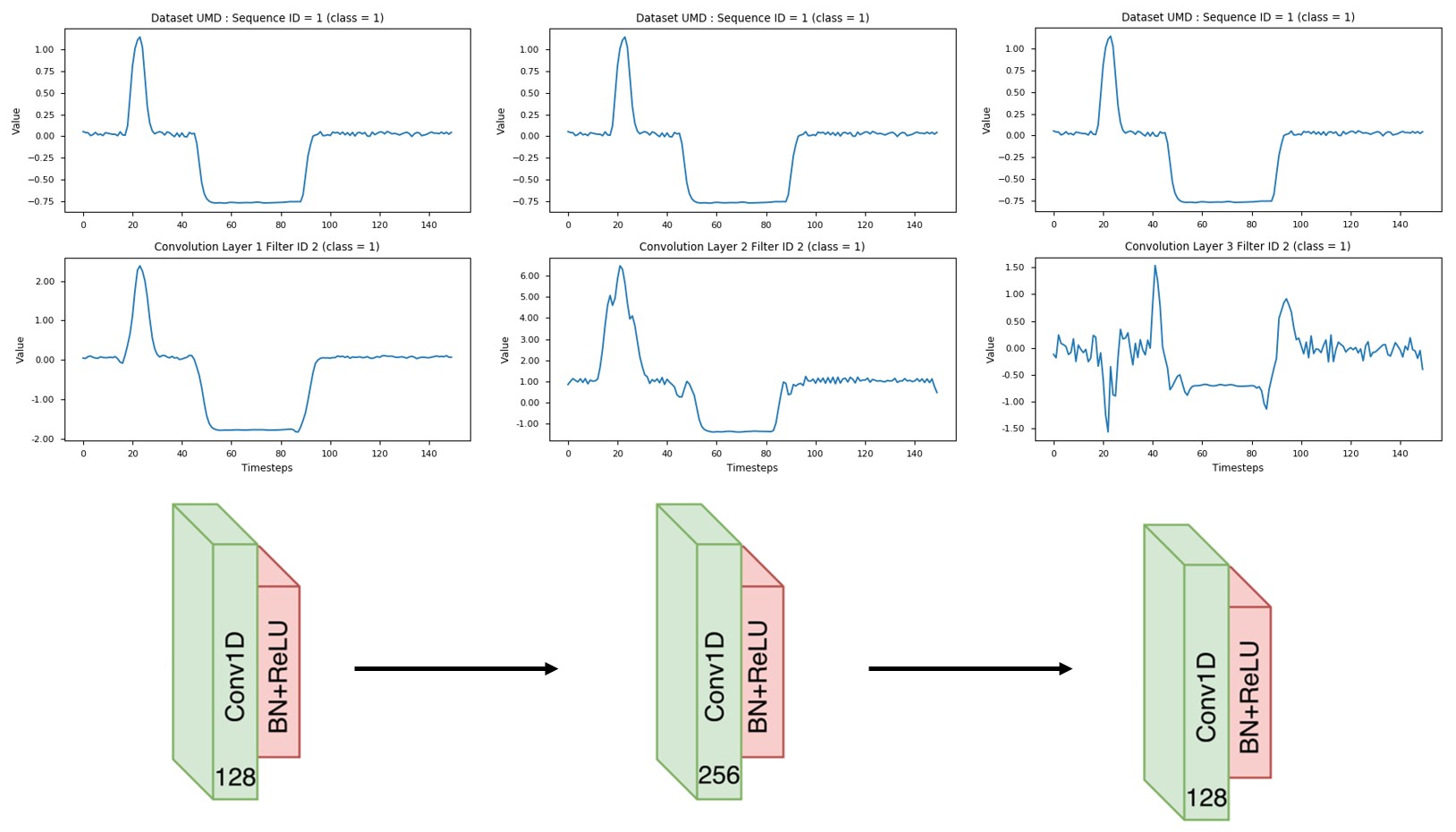}
    \caption{Ablation Test $-$ Visual representation of the input signal after transformation through randomly selected filters from each convolutional layer}
    \label{fig:cnn_ablation}
\end{figure*}

\label{AblationTests}
We perform an ablation study on our model to provide an understanding of the impact of each layer of our model and show how significantly they affect the performance measure. The LSTM-FCN and ALSTM-FCN models are applied to 61 datasets from the UCR repository, such that each dataset is sample \textit{z-normalized}. Each dataset chosen were datasets that outperform the SOTA non-ensemble classifiers, BOSS\cite{Sch_fer_2014} and  WEASEL\cite{schafer2017fast}. We apply BOSS and WEASEL on all UCR datasets based on code and default parameters provided by the author online. It should be noted, this paper is not comparing results with BOSS and WEASEL. BOSS and WEASEL is only used to select datasets that would provide a better understanding of LSTM-FCN and ALSTM-FCN when it performs well.  

In addition, the significance level (alpha) of 0.05 is selected for all statistical tests. The null hypothesis and alternative hypothesis of all Wilcoxon signed-rank test are as follows:
\begin{equation*}
H_o:Median_{\textit{proposed model}} = Median_{\textit{compared model}}
\end{equation*}
\begin{equation*}
H_a:Median_{\textit{proposed model}} \neq Median_{\textit{compared model}}
\end{equation*}.

An essential point of discussion concerning the working of the LSTM-FCN and ALSTM-FCN model is the choice of utilizing an LSTM Recurrent module in conjunction with the FCN block. In the following ablation tests, we study the performance of the individual components which constitute the LSTM-FCN and ALSTM-FCN models, their performance compared to a linear baseline, as well as the empirical and statistical analysis on the performance of the individual components and the final model.

\subsection{Fully Convolutional Block}
LSTM-FCN and ALSTM-FCN comprise of a fully convolutional block and an LSTM/Attention LSTM block. The FCN block has three stacked temporal convolutional blocks with the number of filters defined as 128, 256, and 128. Fig \ref{fig:cnn_ablation} depicts a visual representation of a single sample from the \textit{UMD} dataset after transformation via a random filter selected from each of the convolutional blocks.

As can be noticed, a randomly selected filter from the first CNN block is applying a form of noise reduction that is learned via gradient descent, whereas two subsequent randomly selected filters from the later layers are transforming the data to be far more inconsistent. Based on our analysis of a few filters on various datasets, we conclude that the CNN filters in all layers act as feature extractors and transform the data into separable classes. The model learns the parameters of these transformations on its own via stochastic gradient descent. If a dataset sample requires the removal of noise, it is learned by a few filters of the first CNN layer. It is challenging to postulate what type of transformation is occurring in each filter, as the model transforms the data differently for each of the datasets, on the basis of random initialization of the convolution kernels and order of stochastic gradient descent updates. However, the filter parameters are learned such that their objective is to transform the data into separable classes.

In order to empirically demonstrate that the LSTM-FCN and ALSTM-FCN models are learning to separate the classes better, we examine the features from the FCN block by applying them to a tuned linear SVM classifier. The results are summarized in Table \ref{tab:svm_table}. The linear SVM classifier that is applied on the features extracted from the FCN block is better in 41 datasets (LSTM-FCN model) and 45 datasets (ALSTM-FCN model) as compared to when the tuned linear SVM classifier is applied on to the raw signal. Based on this knowledge, we conclude that the FCN block is transforming the data into separable classes. 

\subsection{LSTM/ALSTM Recurrent Block}
Due to the dimensional shuffle that is applied before the LSTM block, the features extracted by LSTM block by itself do not contribute significantly to the overall performance. When these features are applied onto a tuned linear SVM classifier, the classifier is better in only 19 datasets (for the LSTM block) and 4 datasets (for the ALSTM block) as compared to when the tuned linear SVM classifier is applied to the raw input dataset. The above indicates that the LSTM, by itself, is not separating the data into linear separable classes.  



 \begin{table}[htpb]
  \vspace{-4mm}
 \caption{Ablation Test - Linear SVM performance comparison of LSTM/ALSTM Block, FCN Block with the Raw Signals. Green cells and orange cells designate instances where the linear SVM model on the block exceeds the linear SVM on raw signals. Bold values denotes the block with the best performance using the linear SVM classifier. Count$^*$ represents the number of bold values in that column.} 
 
\label{tab:svm_table}
\begin{adjustbox}{width=1 \linewidth}

 \begin{tabularx}{0.62 \textwidth}{|c|C|C|C|C|C|C|}
    \hline
    Dataset & Raw   & FCN Block & LSTM Block & Raw   & FCN Block & ALSTM Block \\
    \hline
    Car   & \textbf{0.83} & 0.42  & 0.78  & \textbf{0.83} & 0.23  & 0.22 \\
    \hline
    ChlConc & 0.57  & 0.53  & \cellcolor[rgb]{ .663,  .816,  .557}\textbf{0.61} & \textbf{0.57} & 0.53  & 0.53 \\
    \hline
    Compt & 0.52  & 0.50  & \cellcolor[rgb]{ .663,  .816,  .557}\textbf{0.54} & 0.52  & \cellcolor[rgb]{ .929,  .49,  .192}\textbf{0.77} & 0.50 \\
    \hline
    Cricket\_X & 0.27  & \cellcolor[rgb]{ .663,  .816,  .557}\textbf{0.36} & \cellcolor[rgb]{ .663,  .816,  .557}0.33 & 0.27  & \cellcolor[rgb]{ .929,  .49,  .192}\textbf{0.45} & 0.12 \\
    \hline
    Cricket\_Z & 0.28  & \cellcolor[rgb]{ .663,  .816,  .557}0.51 & \cellcolor[rgb]{ .663,  .816,  .557}\textbf{0.54} & 0.28  & \cellcolor[rgb]{ .929,  .49,  .192}\textbf{0.67} & 0.09 \\
    \hline
    DiaSzRed & \textbf{0.94} & 0.31  & 0.93  & \textbf{0.94} & 0.31  & 0.30 \\
    \hline
    DsPhOutAgGp & 0.80  & 0.78  & \cellcolor[rgb]{ .663,  .816,  .557}\textbf{0.81} & 0.80  & \cellcolor[rgb]{ .929,  .49,  .192}\textbf{0.82} & 0.64 \\
    \hline
    DsPhxOCor & 0.53  & \cellcolor[rgb]{ .663,  .816,  .557}\textbf{0.81} & 0.48  & 0.53  & \cellcolor[rgb]{ .929,  .49,  .192}\textbf{0.63} & \cellcolor[rgb]{ .929,  .49,  .192}\textbf{0.63} \\
    \hline
    DsPhxTW & 0.76  & 0.74  & \cellcolor[rgb]{ .663,  .816,  .557}\textbf{0.77} & 0.76  & \cellcolor[rgb]{ .929,  .49,  .192}\textbf{0.79} & 0.53 \\
    \hline
    Earthquakes & 0.57  & \cellcolor[rgb]{ .663,  .816,  .557}\textbf{0.82} & \cellcolor[rgb]{ .663,  .816,  .557}0.71 & 0.57  & \cellcolor[rgb]{ .929,  .49,  .192}\textbf{0.82} & \cellcolor[rgb]{ .929,  .49,  .192}0.78 \\
    \hline
    FaceAll & 0.68  & \cellcolor[rgb]{ .663,  .816,  .557}\textbf{0.92} & \cellcolor[rgb]{ .663,  .816,  .557}0.77 & 0.68  & \cellcolor[rgb]{ .929,  .49,  .192}\textbf{0.95} & 0.18 \\
    \hline
    FordB & 0.49  & \cellcolor[rgb]{ .663,  .816,  .557}\textbf{0.88} & \cellcolor[rgb]{ .663,  .816,  .557}0.56 & 0.49  & 0.49  & \cellcolor[rgb]{ .929,  .49,  .192}\textbf{0.50} \\
    \hline
    Ham   & \textbf{0.70} & 0.51  & 0.66  & \textbf{0.70} & 0.51  & 0.51 \\
    \hline
    Haptics & \textbf{0.44} & 0.19  & 0.41  & \textbf{0.44} & 0.19  & 0.21 \\
    \hline
    ItyPwrDmd & 0.96  & \cellcolor[rgb]{ .663,  .816,  .557}0.97 & \cellcolor[rgb]{ .663,  .816,  .557}\textbf{0.97} & 0.96  & \cellcolor[rgb]{ .929,  .49,  .192}\textbf{0.96} & 0.50 \\
    \hline
    LgKchApp & 0.39  & \cellcolor[rgb]{ .663,  .816,  .557}\textbf{0.52} & \cellcolor[rgb]{ .663,  .816,  .557}0.40 & 0.39  & \cellcolor[rgb]{ .929,  .49,  .192}\textbf{0.68} & 0.33 \\
    \hline
    Lighting7 & 0.64  & \cellcolor[rgb]{ .663,  .816,  .557}\textbf{0.77} & 0.59  & 0.64  & \cellcolor[rgb]{ .929,  .49,  .192}\textbf{0.68} & 0.26 \\
    \hline
    Mallat & \textbf{0.88} & 0.12  & 0.54  & \textbf{0.88} & 0.12  & 0.12 \\
    \hline
    MedImg & 0.56  & \cellcolor[rgb]{ .663,  .816,  .557}\textbf{0.77} & 0.56  & 0.56  & \cellcolor[rgb]{ .929,  .49,  .192}\textbf{0.76} & 0.51 \\
    \hline
    MidPhxOtAgGrp & \textbf{0.80} & 0.75  & 0.80  & \textbf{0.80} & 0.76  & 0.27 \\
    \hline
    MidPhxOtCor & 0.53  & \cellcolor[rgb]{ .663,  .816,  .557}\textbf{0.82} & \cellcolor[rgb]{ .663,  .816,  .557}0.55 & 0.53  & \cellcolor[rgb]{ .929,  .49,  .192}\textbf{0.65} & \cellcolor[rgb]{ .929,  .49,  .192}\textbf{0.65} \\
    \hline
    MidPhxTW & 0.64  & 0.60  & \cellcolor[rgb]{ .663,  .816,  .557}\textbf{0.65} & \textbf{0.64} & 0.61  & 0.21 \\
    \hline
    NonECG\_Thor1 & \textbf{0.91} & 0.19  & 0.85  & \textbf{0.91} & 0.22  & 0.02 \\
    \hline
    NonECG\_Th2 & \textbf{0.92} & 0.17  & 0.20  & \textbf{0.92} & 0.22  & 0.02 \\
    \hline
    OSULeaf & 0.42  & \cellcolor[rgb]{ .663,  .816,  .557}\textbf{0.48} & 0.41  & 0.42  & \cellcolor[rgb]{ .929,  .49,  .192}\textbf{0.50} & 0.18 \\
    \hline
    PhgOtCor & 0.66  & \cellcolor[rgb]{ .663,  .816,  .557}\textbf{0.82} & 0.66  & \textbf{0.66} & 0.61  & 0.61 \\
    \hline
    PrxPhxOtAgeGp & \textbf{0.85} & 0.84  & 0.84  & \textbf{0.85} & 0.84  & 0.49 \\
    \hline
    PrxPhxOtCor & \textbf{0.79} & 0.68  & 0.75  & 0.79  & \cellcolor[rgb]{ .929,  .49,  .192}\textbf{0.91} & 0.68 \\
    \hline
    PrxPhxTW & 0.79  & \cellcolor[rgb]{ .663,  .816,  .557}\textbf{0.81} & 0.65  & 0.79  & \cellcolor[rgb]{ .929,  .49,  .192}\textbf{0.83} & 0.45 \\
    \hline
    ScreenType & \textbf{0.38} & 0.33  & 0.37  & 0.38  & \cellcolor[rgb]{ .929,  .49,  .192}\textbf{0.51} & 0.33 \\
    \hline
    SonyAIBO & 0.66  & \cellcolor[rgb]{ .663,  .816,  .557}\textbf{0.95} & 0.64  & 0.66  & \cellcolor[rgb]{ .929,  .49,  .192}\textbf{0.99} & 0.43 \\
    \hline
    SonyAIBOII & 0.81  & \cellcolor[rgb]{ .663,  .816,  .557}\textbf{0.90} & \cellcolor[rgb]{ .663,  .816,  .557}0.82 & 0.81  & \cellcolor[rgb]{ .929,  .49,  .192}\textbf{0.88} & 0.44 \\
    \hline
    SwdLeaf & 0.79  & \cellcolor[rgb]{ .663,  .816,  .557}\textbf{0.97} & \cellcolor[rgb]{ .663,  .816,  .557}0.81 & 0.79  & \cellcolor[rgb]{ .929,  .49,  .192}\textbf{0.98} & 0.05 \\
    \hline
    Symbols & 0.79  & \cellcolor[rgb]{ .663,  .816,  .557}\textbf{0.92} & \cellcolor[rgb]{ .663,  .816,  .557}0.82 & 0.79  & \cellcolor[rgb]{ .929,  .49,  .192}\textbf{0.81} & 0.17 \\
    \hline
    ToeSeg1 & 0.56  & \cellcolor[rgb]{ .663,  .816,  .557}\textbf{0.97} & \cellcolor[rgb]{ .663,  .816,  .557}0.56 & 0.56  & \cellcolor[rgb]{ .929,  .49,  .192}\textbf{0.86} & 0.55 \\
    \hline
    TwoLeadECG & 0.89  & \cellcolor[rgb]{ .663,  .816,  .557}\textbf{0.99} & 0.64  & 0.89  & \cellcolor[rgb]{ .929,  .49,  .192}\textbf{1.00} & 0.50 \\
    \hline
    ACSF1 & 0.59  & \cellcolor[rgb]{ .663,  .816,  .557}\textbf{0.92} & 0.33  & 0.59  & \cellcolor[rgb]{ .929,  .49,  .192}\textbf{0.88} & 0.10 \\
    \hline
    AllGestWiX & 0.27  & \cellcolor[rgb]{ .663,  .816,  .557}\textbf{0.67} & \cellcolor[rgb]{ .663,  .816,  .557}0.30 & 0.27  & \cellcolor[rgb]{ .929,  .49,  .192}\textbf{0.66} & 0.10 \\
    \hline
    AllGestWiY & 0.35  & \cellcolor[rgb]{ .663,  .816,  .557}\textbf{0.74} & 0.30  & 0.35  & \cellcolor[rgb]{ .929,  .49,  .192}\textbf{0.73} & 0.11 \\
    \hline
    AllGestWiZ & 0.30  & \cellcolor[rgb]{ .663,  .816,  .557}\textbf{0.65} & 0.24  & 0.30  & \cellcolor[rgb]{ .929,  .49,  .192}\textbf{0.65} & 0.10 \\
    \hline
    Chinatown & 0.97  & \cellcolor[rgb]{ .663,  .816,  .557}\textbf{0.98} & 0.91  & 0.97  & \cellcolor[rgb]{ .929,  .49,  .192}\textbf{0.98} & 0.72 \\
    \hline
    Crop  & 0.69  & \cellcolor[rgb]{ .663,  .816,  .557}\textbf{0.72} & 0.45  & 0.69  & \cellcolor[rgb]{ .929,  .49,  .192}\textbf{0.72} & 0.34 \\
    \hline
    EOGHzSgn & 0.43  & \cellcolor[rgb]{ .663,  .816,  .557}\textbf{0.53} & 0.24  & 0.43  & \cellcolor[rgb]{ .929,  .49,  .192}\textbf{0.55} & 0.08 \\
    \hline
    EOGVtSgn & 0.35  & \cellcolor[rgb]{ .663,  .816,  .557}\textbf{0.42} & 0.29  & 0.35  & \cellcolor[rgb]{ .929,  .49,  .192}\textbf{0.35} & 0.08 \\
    \hline
    EthLevel & \textbf{0.75} & 0.75  & 0.25  & \textbf{0.75} & 0.71  & 0.25 \\
    \hline
    FrzRegTr & 0.98  & \cellcolor[rgb]{ .663,  .816,  .557}\textbf{1.00} & 0.80  & 0.98  & \cellcolor[rgb]{ .929,  .49,  .192}\textbf{1.00} & 0.50 \\
    \hline
    GestPebZ1 & 0.72  & \cellcolor[rgb]{ .663,  .816,  .557}\textbf{0.85} & 0.69  & 0.72  & \cellcolor[rgb]{ .929,  .49,  .192}\textbf{0.87} & 0.22 \\
    \hline
    GunPointMVsF & 0.98  & \cellcolor[rgb]{ .663,  .816,  .557}\textbf{1.00} & 0.89  & 0.98  & \cellcolor[rgb]{ .929,  .49,  .192}\textbf{1.00} & 0.53 \\
    \hline
    GunPointOVsY & 0.88  & \cellcolor[rgb]{ .663,  .816,  .557}\textbf{0.97} & 0.86  & 0.88  & \cellcolor[rgb]{ .929,  .49,  .192}\textbf{0.97} & 0.48 \\
    \hline
    InsEPGRegTr & 0.64  & \cellcolor[rgb]{ .663,  .816,  .557}\textbf{1.00} & 0.59  & 0.64  & \cellcolor[rgb]{ .929,  .49,  .192}\textbf{1.00} & 0.47 \\
    \hline
    MelbPed & 0.84  & \cellcolor[rgb]{ .663,  .816,  .557}\textbf{0.91} & 0.75  & 0.84  & \cellcolor[rgb]{ .929,  .49,  .192}\textbf{0.90} & 0.15 \\
    \hline
    MxShpRegTr & 0.81  & \cellcolor[rgb]{ .663,  .816,  .557}\textbf{0.95} & 0.80  & 0.81  & \cellcolor[rgb]{ .929,  .49,  .192}\textbf{0.96} & 0.17 \\
    \hline
    MxShpSmlTr & 0.80  & \cellcolor[rgb]{ .663,  .816,  .557}\textbf{0.91} & \cellcolor[rgb]{ .663,  .816,  .557}0.81 & 0.80  & \cellcolor[rgb]{ .929,  .49,  .192}\textbf{0.92} & 0.13 \\
    \hline
    PickGestWiZ & 0.60  & \cellcolor[rgb]{ .663,  .816,  .557}\textbf{0.70} & 0.50  & 0.60  & \cellcolor[rgb]{ .929,  .49,  .192}\textbf{0.68} & 0.10 \\
    \hline
    PigAryPress & 0.06  & \cellcolor[rgb]{ .663,  .816,  .557}\textbf{0.35} & 0.02  & 0.06  & \cellcolor[rgb]{ .929,  .49,  .192}\textbf{0.37} & 0.02 \\
    \hline
    PowerCons & \textbf{0.93} & 0.89  & 0.83  & \textbf{0.93} & 0.87  & 0.50 \\
    \hline
    SgHdGendCh2 & \textbf{0.88} & 0.82  & 0.81  & \textbf{0.88} & 0.77  & 0.35 \\
    \hline
    SgHdMovCh2 & 0.48  & \cellcolor[rgb]{ .663,  .816,  .557}\textbf{0.51} & 0.35  & 0.48  & \cellcolor[rgb]{ .929,  .49,  .192}\textbf{0.51} & 0.17 \\
    \hline
    ShkGestWiZ & 0.62  & \cellcolor[rgb]{ .663,  .816,  .557}\textbf{0.86} & 0.36  & 0.62  & \cellcolor[rgb]{ .929,  .49,  .192}\textbf{0.86} & 0.10 \\
    \hline
    SmthSub & 0.67  & \cellcolor[rgb]{ .663,  .816,  .557}\textbf{0.97} & 0.61  & 0.67  & \cellcolor[rgb]{ .929,  .49,  .192}\textbf{0.96} & 0.33 \\
    \hline
    UMD   & \textbf{0.98} & 0.97  & 0.71  & 0.98  & \cellcolor[rgb]{ .929,  .49,  .192}\textbf{0.99} & 0.22 \\
    \hline
    Count & 15    & 38    & 7     & 15    & 46    & 3 \\
    \hline

    \end{tabularx}%
    \end{adjustbox}
  \label{tab:svm_table}%

\end{table}%

\subsection{LSTM/ALSTM Concatenated With FCN Block} \label{concatlayers}
Nevertheless, when the features of the LSTM block/ALSTM block are concatenated with the CNN features, we obtain a more robust set of features that can better separate the classes of the dataset. The above insight is statistically validated by applying the concatenated features to a single layer perceptron classifier which accepts the extracted features as input (due to the fact that the data is transformed into separable classes). The training scheme of all perceptron models is kept consistent with how we train all LSTM-FCN and ALSTM-FCN models, as detailed in Section \ref{Experiments}. Results, shown in Table \ref{tab:mlp_table}, show that the features from of the LSTM/ALSTM block coupled with the features from the FCN block improve the model performance.




 \begin{table}[htpb]
 \vspace{-4mm}
 \centering
 \caption{Ablation Test - MLP performance comparison of LSTM/ALSTM Block, FCN Block, LSTM/ALSTM-FCN Block and the Raw Signals. Green cells and orange cells designate instances where the MLP model on the block exceeds the MLP on raw signals. Bold values denotes the block with the best performance using the MLP classifier. Count$^{*}$ represents the number of bold values in that column.}
\label{tab:mlp_table}
\begin{adjustbox}{width=1 \linewidth}

 \begin{tabularx}{0.62 \textwidth}{|c|C|C|C|C|C|C|C|C|}

    \hline
    Dataset & Raw   & FCN Block & LSTM Block & LSTM-FCN & Raw   & FCN Block & ALSTM Block & ALSTM-FCN \\
    \hline
    Car   & 0.83  & 0.42  & 0.78  & \cellcolor[rgb]{ .663,  .816,  .557}\textbf{0.95} & 0.83  & 0.23  & 0.22  & \cellcolor[rgb]{ .929,  .49,  .192}\textbf{0.92} \\
    \hline
    ChlConc & 0.57  & 0.53  & \cellcolor[rgb]{ .663,  .816,  .557}0.61 & \cellcolor[rgb]{ .663,  .816,  .557}\textbf{0.80} & 0.57  & 0.53  & 0.53  & \cellcolor[rgb]{ .929,  .49,  .192}\textbf{0.79} \\
    \hline
    Compt & 0.52  & 0.50  & \cellcolor[rgb]{ .663,  .816,  .557}0.54 & \cellcolor[rgb]{ .663,  .816,  .557}\textbf{0.84} & 0.52  & \cellcolor[rgb]{ .929,  .49,  .192}0.77 & 0.50  & \cellcolor[rgb]{ .929,  .49,  .192}\textbf{0.84} \\
    \hline
    Cricket\_X & 0.27  & \cellcolor[rgb]{ .663,  .816,  .557}0.36 & \cellcolor[rgb]{ .663,  .816,  .557}0.33 & \cellcolor[rgb]{ .663,  .816,  .557}\textbf{0.78} & 0.27  & \cellcolor[rgb]{ .929,  .49,  .192}0.45 & 0.12  & \cellcolor[rgb]{ .929,  .49,  .192}\textbf{0.78} \\
    \hline
    Cricket\_Z & 0.28  & \cellcolor[rgb]{ .663,  .816,  .557}0.51 & \cellcolor[rgb]{ .663,  .816,  .557}0.54 & \cellcolor[rgb]{ .663,  .816,  .557}\textbf{0.82} & 0.28  & \cellcolor[rgb]{ .929,  .49,  .192}0.67 & 0.09  & \cellcolor[rgb]{ .929,  .49,  .192}\textbf{0.79} \\
    \hline
    DiaSzRed & \textbf{0.94} & 0.31  & 0.93  & \textbf{0.94} & \textbf{0.94} & 0.31  & 0.30  & 0.94 \\
    \hline
    DsPhOutAgGp & 0.80  & 0.78  & \cellcolor[rgb]{ .663,  .816,  .557}0.81 & \cellcolor[rgb]{ .663,  .816,  .557}\textbf{0.83} & 0.80  & \cellcolor[rgb]{ .929,  .49,  .192}0.82 & 0.64  & \cellcolor[rgb]{ .929,  .49,  .192}\textbf{0.83} \\
    \hline
    DsPhxOCor & 0.53  & \cellcolor[rgb]{ .663,  .816,  .557}0.81 & 0.48  & \cellcolor[rgb]{ .663,  .816,  .557}\textbf{0.81} & 0.53  & \cellcolor[rgb]{ .929,  .49,  .192}0.63 & \cellcolor[rgb]{ .929,  .49,  .192}0.63 & \cellcolor[rgb]{ .929,  .49,  .192}\textbf{0.81} \\
    \hline
    DsPhxTW & 0.76  & 0.74  & \cellcolor[rgb]{ .663,  .816,  .557}0.77 & \cellcolor[rgb]{ .663,  .816,  .557}\textbf{0.79} & 0.76  & \cellcolor[rgb]{ .929,  .49,  .192}0.79 & 0.53  & \cellcolor[rgb]{ .929,  .49,  .192}\textbf{0.79} \\
    \hline
    Earthquakes & 0.57  & \cellcolor[rgb]{ .663,  .816,  .557}\textbf{0.82} & \cellcolor[rgb]{ .663,  .816,  .557}0.71 & \cellcolor[rgb]{ .663,  .816,  .557}0.80 & 0.57  & \cellcolor[rgb]{ .929,  .49,  .192}\textbf{0.82} & \cellcolor[rgb]{ .929,  .49,  .192}0.78 & \cellcolor[rgb]{ .929,  .49,  .192}0.80 \\
    \hline
    FaceAll & 0.68  & \cellcolor[rgb]{ .663,  .816,  .557}0.92 & \cellcolor[rgb]{ .663,  .816,  .557}0.77 & \cellcolor[rgb]{ .663,  .816,  .557}\textbf{0.92} & 0.68  & \cellcolor[rgb]{ .929,  .49,  .192}\textbf{0.95} & 0.18  & \cellcolor[rgb]{ .929,  .49,  .192}0.93 \\
    \hline
    FordB & 0.49  & \cellcolor[rgb]{ .663,  .816,  .557}0.88 & \cellcolor[rgb]{ .663,  .816,  .557}0.56 & \cellcolor[rgb]{ .663,  .816,  .557}\textbf{0.89} & 0.49  & 0.49  & \cellcolor[rgb]{ .929,  .49,  .192}0.50 & \cellcolor[rgb]{ .929,  .49,  .192}\textbf{0.88} \\
    \hline
    Ham   & 0.70  & 0.51  & 0.66  & \cellcolor[rgb]{ .663,  .816,  .557}\textbf{0.73} & 0.70  & 0.51  & 0.51  & \cellcolor[rgb]{ .929,  .49,  .192}\textbf{0.72} \\
    \hline
    Haptics & 0.44  & 0.19  & 0.41  & \cellcolor[rgb]{ .663,  .816,  .557}\textbf{0.49} & 0.44  & 0.19  & 0.21  & \cellcolor[rgb]{ .929,  .49,  .192}\textbf{0.50} \\
    \hline
    ItyPwrDmd & 0.96  & \cellcolor[rgb]{ .663,  .816,  .557}0.97 & \cellcolor[rgb]{ .663,  .816,  .557}\textbf{0.97} & \cellcolor[rgb]{ .663,  .816,  .557}0.96 & 0.96  & \cellcolor[rgb]{ .929,  .49,  .192}\textbf{0.96} & 0.50  & \cellcolor[rgb]{ .929,  .49,  .192}\textbf{0.96} \\
    \hline
    LgKchApp & 0.39  & \cellcolor[rgb]{ .663,  .816,  .557}0.52 & \cellcolor[rgb]{ .663,  .816,  .557}0.40 & \cellcolor[rgb]{ .663,  .816,  .557}\textbf{0.89} & 0.39  & \cellcolor[rgb]{ .929,  .49,  .192}0.68 & 0.33  & \cellcolor[rgb]{ .929,  .49,  .192}\textbf{0.91} \\
    \hline
    Lighting7 & 0.64  & \cellcolor[rgb]{ .663,  .816,  .557}0.77 & 0.59  & \cellcolor[rgb]{ .663,  .816,  .557}\textbf{0.79} & 0.64  & \cellcolor[rgb]{ .929,  .49,  .192}0.68 & 0.26  & \cellcolor[rgb]{ .929,  .49,  .192}\textbf{0.77} \\
    \hline
    Mallat & 0.88  & 0.12  & 0.54  & \cellcolor[rgb]{ .663,  .816,  .557}\textbf{0.97} & 0.88  & 0.12  & 0.12  & \cellcolor[rgb]{ .929,  .49,  .192}\textbf{0.97} \\
    \hline
    MedImg & 0.56  & \cellcolor[rgb]{ .663,  .816,  .557}0.77 & 0.56  & \cellcolor[rgb]{ .663,  .816,  .557}\textbf{0.79} & 0.56  & \cellcolor[rgb]{ .929,  .49,  .192}0.76 & 0.51  & \cellcolor[rgb]{ .929,  .49,  .192}\textbf{0.77} \\
    \hline
    MidPhxOtAgGrp & \textbf{0.80} & 0.75  & 0.80  & 0.75  & \textbf{0.80} & 0.76  & 0.27  & 0.76 \\
    \hline
    MidPhxOtCor & 0.53  & \cellcolor[rgb]{ .663,  .816,  .557}0.82 & \cellcolor[rgb]{ .663,  .816,  .557}0.55 & \cellcolor[rgb]{ .663,  .816,  .557}\textbf{0.83} & 0.53  & \cellcolor[rgb]{ .929,  .49,  .192}0.65 & \cellcolor[rgb]{ .929,  .49,  .192}0.65 & \cellcolor[rgb]{ .929,  .49,  .192}\textbf{0.83} \\
    \hline
    MidPhxTW & 0.64  & 0.60  & \cellcolor[rgb]{ .663,  .816,  .557}\textbf{0.65} & 0.61  & \textbf{0.64} & 0.61  & 0.21  & 0.61 \\
    \hline
    NonECG\_Thor1 & 0.91  & 0.19  & 0.85  & \cellcolor[rgb]{ .663,  .816,  .557}\textbf{0.96} & 0.91  & 0.22  & 0.02  & \cellcolor[rgb]{ .929,  .49,  .192}\textbf{0.96} \\
    \hline
    NonECG\_Th2 & 0.92  & 0.17  & 0.20  & \cellcolor[rgb]{ .663,  .816,  .557}\textbf{0.96} & 0.92  & 0.22  & 0.02  & \cellcolor[rgb]{ .929,  .49,  .192}\textbf{0.95} \\
    \hline
    OSULeaf & 0.42  & \cellcolor[rgb]{ .663,  .816,  .557}0.48 & 0.41  & \cellcolor[rgb]{ .663,  .816,  .557}\textbf{0.98} & 0.42  & \cellcolor[rgb]{ .929,  .49,  .192}0.50 & 0.18  & \cellcolor[rgb]{ .929,  .49,  .192}\textbf{0.99} \\
    \hline
    PhgOtCor & 0.66  & \cellcolor[rgb]{ .663,  .816,  .557}0.82 & 0.66  & \cellcolor[rgb]{ .663,  .816,  .557}\textbf{0.83} & 0.66  & 0.61  & 0.61  & \cellcolor[rgb]{ .929,  .49,  .192}\textbf{0.83} \\
    \hline
    PrxPhxOtAgeGp & 0.85  & 0.84  & 0.84  & \cellcolor[rgb]{ .663,  .816,  .557}\textbf{0.87} & 0.85  & 0.84  & 0.49  & \cellcolor[rgb]{ .929,  .49,  .192}\textbf{0.86} \\
    \hline
    PrxPhxOtCor & 0.79  & 0.68  & 0.75  & \cellcolor[rgb]{ .663,  .816,  .557}\textbf{0.91} & 0.79  & \cellcolor[rgb]{ .929,  .49,  .192}0.91 & 0.68  & \cellcolor[rgb]{ .929,  .49,  .192}\textbf{0.92} \\
    \hline
    PrxPhxTW & 0.79  & \cellcolor[rgb]{ .663,  .816,  .557}0.81 & 0.65  & \cellcolor[rgb]{ .663,  .816,  .557}\textbf{0.82} & 0.79  & \cellcolor[rgb]{ .929,  .49,  .192}\textbf{0.83} & 0.45  & \cellcolor[rgb]{ .929,  .49,  .192}0.82 \\
    \hline
    ScreenType & 0.38  & 0.33  & 0.37  & \cellcolor[rgb]{ .663,  .816,  .557}\textbf{0.62} & 0.38  & \cellcolor[rgb]{ .929,  .49,  .192}0.51 & 0.33  & \cellcolor[rgb]{ .929,  .49,  .192}\textbf{0.64} \\
    \hline
    SonyAIBO & 0.66  & \cellcolor[rgb]{ .663,  .816,  .557}0.95 & 0.64  & \cellcolor[rgb]{ .663,  .816,  .557}\textbf{0.96} & 0.66  & \cellcolor[rgb]{ .929,  .49,  .192}\textbf{0.99} & 0.43  & \cellcolor[rgb]{ .929,  .49,  .192}0.97 \\
    \hline
    SonyAIBOII & 0.81  & \cellcolor[rgb]{ .663,  .816,  .557}0.90 & \cellcolor[rgb]{ .663,  .816,  .557}0.82 & \cellcolor[rgb]{ .663,  .816,  .557}\textbf{0.97} & 0.81  & \cellcolor[rgb]{ .929,  .49,  .192}0.88 & 0.44  & \cellcolor[rgb]{ .929,  .49,  .192}\textbf{0.98} \\
    \hline
    SwdLeaf & 0.79  & \cellcolor[rgb]{ .663,  .816,  .557}0.97 & \cellcolor[rgb]{ .663,  .816,  .557}0.81 & \cellcolor[rgb]{ .663,  .816,  .557}\textbf{0.98} & 0.79  & \cellcolor[rgb]{ .929,  .49,  .192}\textbf{0.98} & 0.05  & \cellcolor[rgb]{ .929,  .49,  .192}0.97 \\
    \hline
    Symbols & 0.79  & \cellcolor[rgb]{ .663,  .816,  .557}0.92 & \cellcolor[rgb]{ .663,  .816,  .557}0.82 & \cellcolor[rgb]{ .663,  .816,  .557}\textbf{0.98} & 0.79  & \cellcolor[rgb]{ .929,  .49,  .192}0.81 & 0.17  & \cellcolor[rgb]{ .929,  .49,  .192}\textbf{0.97} \\
    \hline
    ToeSeg1 & 0.56  & \cellcolor[rgb]{ .663,  .816,  .557}0.97 & \cellcolor[rgb]{ .663,  .816,  .557}0.56 & \cellcolor[rgb]{ .663,  .816,  .557}\textbf{0.98} & 0.56  & \cellcolor[rgb]{ .929,  .49,  .192}0.86 & 0.55  & \cellcolor[rgb]{ .929,  .49,  .192}\textbf{0.99} \\
    \hline
    TwoLeadECG & 0.89  & \cellcolor[rgb]{ .663,  .816,  .557}0.99 & 0.64  & \cellcolor[rgb]{ .663,  .816,  .557}\textbf{1.00} & 0.89  & \cellcolor[rgb]{ .929,  .49,  .192}1.00 & 0.50  & \cellcolor[rgb]{ .929,  .49,  .192}\textbf{1.00} \\
    \hline
    ACSF1 & 0.40  & \cellcolor[rgb]{ .663,  .816,  .557}\textbf{0.91} & 0.36  & \cellcolor[rgb]{ .663,  .816,  .557}\textbf{0.91} & 0.40  & \cellcolor[rgb]{ .929,  .49,  .192}0.89 & 0.10  & \cellcolor[rgb]{ .929,  .49,  .192}\textbf{0.90} \\
    \hline
    AllGestWiX & 0.24  & \cellcolor[rgb]{ .663,  .816,  .557}0.66 & \cellcolor[rgb]{ .663,  .816,  .557}0.30 & \cellcolor[rgb]{ .663,  .816,  .557}\textbf{0.71} & 0.24  & \cellcolor[rgb]{ .929,  .49,  .192}0.65 & 0.10  & \cellcolor[rgb]{ .929,  .49,  .192}\textbf{0.71} \\
    \hline
    AllGestWiY & 0.32  & \cellcolor[rgb]{ .663,  .816,  .557}0.72 & 0.29  & \cellcolor[rgb]{ .663,  .816,  .557}\textbf{0.79} & 0.32  & \cellcolor[rgb]{ .929,  .49,  .192}0.73 & 0.10  & \cellcolor[rgb]{ .929,  .49,  .192}\textbf{0.77} \\
    \hline
    AllGestWiZ & 0.26  & \cellcolor[rgb]{ .663,  .816,  .557}0.63 & 0.25  & \cellcolor[rgb]{ .663,  .816,  .557}\textbf{0.68} & 0.25  & \cellcolor[rgb]{ .929,  .49,  .192}0.64 & 0.10  & \cellcolor[rgb]{ .929,  .49,  .192}\textbf{0.68} \\
    \hline
    Chinatown & 0.96  & \cellcolor[rgb]{ .663,  .816,  .557}\textbf{0.98} & 0.94  & \cellcolor[rgb]{ .663,  .816,  .557}\textbf{0.98} & 0.96  & \cellcolor[rgb]{ .929,  .49,  .192}\textbf{0.98} & 0.72  & \cellcolor[rgb]{ .929,  .49,  .192}\textbf{0.98} \\
    \hline
    Crop  & 0.63  & \cellcolor[rgb]{ .663,  .816,  .557}0.72 & 0.42  & \cellcolor[rgb]{ .663,  .816,  .557}\textbf{0.74} & 0.63  & \cellcolor[rgb]{ .929,  .49,  .192}0.71 & 0.34  & \cellcolor[rgb]{ .929,  .49,  .192}\textbf{0.74} \\
    \hline
    EOGHzSgn & 0.30  & \cellcolor[rgb]{ .663,  .816,  .557}0.54 & 0.27  & \cellcolor[rgb]{ .663,  .816,  .557}\textbf{0.62} & 0.30  & \cellcolor[rgb]{ .929,  .49,  .192}0.55 & 0.08  & \cellcolor[rgb]{ .929,  .49,  .192}\textbf{0.60} \\
    \hline
    EOGVtSgn & 0.32  & \cellcolor[rgb]{ .663,  .816,  .557}0.36 & 0.28  & \cellcolor[rgb]{ .663,  .816,  .557}\textbf{0.49} & 0.31  & \cellcolor[rgb]{ .929,  .49,  .192}0.35 & 0.08  & \cellcolor[rgb]{ .929,  .49,  .192}\textbf{0.45} \\
    \hline
    EthLevel & 0.46  & \cellcolor[rgb]{ .663,  .816,  .557}0.62 & 0.25  & \cellcolor[rgb]{ .663,  .816,  .557}\textbf{0.68} & 0.53  & \cellcolor[rgb]{ .929,  .49,  .192}0.65 & 0.25  & \cellcolor[rgb]{ .929,  .49,  .192}\textbf{0.73} \\
    \hline
    FrzRegTr & 0.82  & \cellcolor[rgb]{ .663,  .816,  .557}1.00 & 0.78  & \cellcolor[rgb]{ .663,  .816,  .557}\textbf{1.00} & 0.80  & \cellcolor[rgb]{ .929,  .49,  .192}1.00 & 0.50  & \cellcolor[rgb]{ .929,  .49,  .192}\textbf{1.00} \\
    \hline
    GestPebZ1 & 0.73  & \cellcolor[rgb]{ .663,  .816,  .557}0.81 & 0.70  & \cellcolor[rgb]{ .663,  .816,  .557}\textbf{0.94} & 0.71  & \cellcolor[rgb]{ .929,  .49,  .192}0.84 & 0.18  & \cellcolor[rgb]{ .929,  .49,  .192}\textbf{0.91} \\
    \hline
    GunPointMVsF & 0.85  & \cellcolor[rgb]{ .663,  .816,  .557}1.00 & \cellcolor[rgb]{ .663,  .816,  .557}0.88 & \cellcolor[rgb]{ .663,  .816,  .557}\textbf{1.00} & 0.84  & \cellcolor[rgb]{ .929,  .49,  .192}1.00 & 0.53  & \cellcolor[rgb]{ .929,  .49,  .192}\textbf{1.00} \\
    \hline
    GunPointOVsY & 0.88  & \cellcolor[rgb]{ .663,  .816,  .557}0.97 & 0.84  & \cellcolor[rgb]{ .663,  .816,  .557}\textbf{1.00} & 0.85  & \cellcolor[rgb]{ .929,  .49,  .192}0.97 & 0.48  & \cellcolor[rgb]{ .929,  .49,  .192}\textbf{0.99} \\
    \hline
    InsEPGRegTr & 0.65  & \cellcolor[rgb]{ .663,  .816,  .557}0.98 & 0.62  & \cellcolor[rgb]{ .663,  .816,  .557}\textbf{1.00} & 0.65  & \cellcolor[rgb]{ .929,  .49,  .192}0.98 & 0.47  & \cellcolor[rgb]{ .929,  .49,  .192}\textbf{1.00} \\
    \hline
    MelbPed & 0.79  & \cellcolor[rgb]{ .663,  .816,  .557}0.91 & 0.74  & \cellcolor[rgb]{ .663,  .816,  .557}\textbf{0.91} & 0.78  & \cellcolor[rgb]{ .929,  .49,  .192}0.90 & 0.13  & \cellcolor[rgb]{ .929,  .49,  .192}\textbf{0.91} \\
    \hline
    MxShpRegTr & 0.80  & \cellcolor[rgb]{ .663,  .816,  .557}0.95 & 0.80  & \cellcolor[rgb]{ .663,  .816,  .557}\textbf{0.97} & 0.81  & \cellcolor[rgb]{ .929,  .49,  .192}0.96 & 0.24  & \cellcolor[rgb]{ .929,  .49,  .192}\textbf{0.97} \\
    \hline
    MxShpSmlTr & 0.75  & \cellcolor[rgb]{ .663,  .816,  .557}0.92 & \cellcolor[rgb]{ .663,  .816,  .557}0.80 & \cellcolor[rgb]{ .663,  .816,  .557}\textbf{0.94} & 0.75  & \cellcolor[rgb]{ .929,  .49,  .192}0.92 & 0.19  & \cellcolor[rgb]{ .929,  .49,  .192}\textbf{0.92} \\
    \hline
    PickGestWiZ & 0.52  & \cellcolor[rgb]{ .663,  .816,  .557}\textbf{0.70} & 0.52  & \cellcolor[rgb]{ .663,  .816,  .557}\textbf{0.70} & 0.52  & \cellcolor[rgb]{ .929,  .49,  .192}0.58 & 0.10  & \cellcolor[rgb]{ .929,  .49,  .192}\textbf{0.64} \\
    \hline
    PigAryPress & 0.05  & \cellcolor[rgb]{ .663,  .816,  .557}0.39 & 0.01  & \cellcolor[rgb]{ .663,  .816,  .557}\textbf{0.42} & 0.05  & \cellcolor[rgb]{ .929,  .49,  .192}\textbf{0.38} & 0.02  & \cellcolor[rgb]{ .929,  .49,  .192}0.38 \\
    \hline
    PowerCons & 0.89  & 0.87  & 0.82  & \cellcolor[rgb]{ .663,  .816,  .557}\textbf{0.93} & 0.89  & 0.88  & 0.50  & \cellcolor[rgb]{ .929,  .49,  .192}\textbf{0.91} \\
    \hline
    SgHdGendCh2 & 0.75  & \cellcolor[rgb]{ .663,  .816,  .557}0.81 & \cellcolor[rgb]{ .663,  .816,  .557}0.80 & \cellcolor[rgb]{ .663,  .816,  .557}\textbf{0.92} & 0.76  & \cellcolor[rgb]{ .929,  .49,  .192}0.79 & 0.65  & \cellcolor[rgb]{ .929,  .49,  .192}\textbf{0.86} \\
    \hline
    SgHdMovCh2 & 0.38  & \cellcolor[rgb]{ .663,  .816,  .557}0.46 & 0.34  & \cellcolor[rgb]{ .663,  .816,  .557}\textbf{0.54} & 0.38  & \cellcolor[rgb]{ .929,  .49,  .192}0.48 & 0.17  & \cellcolor[rgb]{ .929,  .49,  .192}\textbf{0.56} \\
    \hline
    ShkGestWiZ & 0.44  & \cellcolor[rgb]{ .663,  .816,  .557}0.86 & 0.30  & \cellcolor[rgb]{ .663,  .816,  .557}\textbf{0.90} & 0.44  & \cellcolor[rgb]{ .929,  .49,  .192}\textbf{0.88} & 0.10  & \cellcolor[rgb]{ .929,  .49,  .192}\textbf{0.88} \\
    \hline
    SmthSub & 0.69  & \cellcolor[rgb]{ .663,  .816,  .557}0.97 & 0.57  & \cellcolor[rgb]{ .663,  .816,  .557}\textbf{0.99} & 0.70  & \cellcolor[rgb]{ .929,  .49,  .192}0.95 & 0.33  & \cellcolor[rgb]{ .929,  .49,  .192}\textbf{0.98} \\
    \hline
    UMD   & 0.88  & \cellcolor[rgb]{ .663,  .816,  .557}0.94 & 0.73  & \cellcolor[rgb]{ .663,  .816,  .557}\textbf{0.97} & 0.88  & \cellcolor[rgb]{ .929,  .49,  .192}\textbf{0.98} & 0.22  & \cellcolor[rgb]{ .929,  .49,  .192}0.97 \\
    \hline
    Count & 2     & 4     & 2     & 57    & 3     & 10    & 0     & 51 \\
    \hline

    \end{tabularx}%
    \end{adjustbox}
  \label{tab:mlp_table}%

\end{table}%

For the ALSTM-FCN model, the ALSTM features joined with the FCN features outperform the features from the ALSTM block or the FCN block on 49 datasets, yielding to a p-value of 1.34\textit{e-08} when a Wilcoxon Signed-rank test \cite{wilcoxon1964some} is applied. Similarly, the LSTM features joined with the FCN features in the model LSTM-FCN outperform the features from the LSTM block or the FCN block on 54 datasets, yielding to a p-value of 1.22\textit{e-08}. The \textit{Dunn-Sidak} \cite{abdi2007bonferroni} corrected significant alpha value is 0.02. 

It is evident that when applying the LSTM block (with dimension shuffle) and the FCN block parallelly, the blocks augment each other, and force each other to detect a set of features which when combined, yield an overall better performing model. In other words, the LSTM block attached with the FCN block statistically helps improve the overall performance of the model providing informative features that in conjunction with the FCN features, are useful in separating the classes further.

\subsection{Dimension Shuffle vs No Dimension Shuffle}
Another ablation test performed is to check the impact dimension shuffle has on the overall behavior of the model. The dimension shuffle transposes the input univariate time series of $N$ time steps and $1$ variable into a multivariate time series of $N$ variables and $1$ time step. In other words, when dimension shuffle is applied to the input before the LSTM block, the LSTM block will process only $1$ time step with $N$ variables. 

In this ablation test, LSTM-FCN with dimension shuffle is compared to LSTM-FCN without dimension shuffle on all 128 UCR datasets using a cell size of 8, 64, 128 (yielding to a total of $128 \times 3 = 384$ experiments). LSTM-FCN with dimension shuffle outperforms LSTM-FCN without dimension shuffle on 258 experiments, ties in 27 experiments, and performs worse in 99 experiments. For the experiments when LSTM-FCN with dimension shuffle outperforms LSTM-FCN without dimension shuffle, the accuracy improved on average by $6.00 \%$. Conversely, for the experiments when LSTM-FCN with dimension shuffle performs worse than LSTM-FCN without dimension shuffle, the accurcy is worse by an average of $5.26 \%$. A Wilcoxson signed-rank test results in a p-value of $3.69E-17$, indicating a statistical difference in performance where LSTM-FCN with dimension shuffle performs better. This result is contrary to what most people would hypothesize. LSTM-FCN without dimension shuffle overfits the UCR datasets in more instances than LSTM-FCN with dimension shuffle. This is because the LSTM block without dimension shuffle by itself performs extremely well. The FCN block and LSTM block without the dimension shuffle does not benefit each other.  

Another critical fact to note is that the LSTM-FCN with dimension shuffle processes the univariate time series in one time step. The gating mechanisms of the LSTM-FCN is only being applied on a single time step. This attributes to why LSTM with dimension shuffle by itself performs poorly. However, as noticed in Section \ref{concatlayers}, when applying the LSTM block with dimension shuffle and the FCN block parallelly, the blocks augment each other, while improving its overall performance. To the best of our knowledge, we believe the LSTM block with a dimension shuffle acts as a regularizer to the FCN block, forcing the FCN block to improve its performance.

\subsection{Replacing LSTM with GRU, RNN, and a Dense Layer}

Since the usage of the LSTM block when applying dimension shuffle to the input is atypical, we replace the LSTM block with a GRU block (8, 64, 128 cells), basic RNN block (8, 64, 128 cells), and a Dense block with a sigmoid activation function (8, 64, 128 units) on all 128 datasets (total of 384 experiments on each model). \hl{The intuition behind selecting an RNN block and a GRU block is that these blocks have similar properties to an LSTM block, and differ only in their capacity to learn long term temporal dependencies. Furthermore, a dense layer is selected to compare against the atypical usage of the LSTM block, so that we may analyze whether the complex interaction within the recurrent gates of the LSTM can be simplified into a single fully connected layer.} We chose the sigmoid activation function for the Dense block, instead of the standard Rectifying Linear Unit (ReLU) activation, as we wish to compare the effectiveness of the gating effect exhibited by the 3 gates of the LSTM. The majority of the gates of the LSTM use the sigmoid activation function. Therefore, we construct the Dense block to also use the same. The input to the GRU block, RNN block, and Dense block had a dimension shuffle applied onto it. Replacing the LSTM block of LSTM-FCN with a GRU block was first proposed by \textit{Elsayed et. al} \cite{elsayed2018deep}. Table \ref{tab:substitute} summarizes a Wilcoxson signed-rank test when LSTM-FCN with dimension shuffle is compared to GRU-FCN, RNN-FCN, and Dense-FCN.

 \newcolumntype{C}{>{\centering\arraybackslash}X}



 \begin{table}[htpb]
 \centering
 \caption{Ablation Test - Wilcoxson signed-rank test comparing LSTM-FCN with GRU-FCN, RNN-FCN, and Dense-FCN. The values in parenthesis depicts the number of wins, ties, and losses the row index has with the header. Red cell depicts when the test fails to reject the null hypothesis.}
\begin{adjustbox}{width=1 \linewidth}

 \begin{tabularx}{0.62 \textwidth}{|C|C|C|C|}
    \hline
          & \textbf{GRU-FCN} & \textbf{RNN-FCN} & \textbf{Dense-FCN}\\
    \hline
    \textbf{LSTM-FCN} & 6.33E-16 (243/38/103) & 1.34E-17 (247/41/96) & 2.81E-10 (231/35/118) \\
    \hline
    \textbf{GRU-FCN} &       & 1.05E-02 (185/53/146) & \cellcolor[rgb]{ 1,  .78,  .808}\textcolor[rgb]{ .612,  0,  .024}{1.57E-01 (160/49/175)} \\
    \hline
    \textbf{RNN-FCN} &       &       & 7.55E-05 (135/49/200) \\
    \hline

    \end{tabularx}%
    \end{adjustbox}
  \label{tab:substitute}%

\end{table}%

The Wilcoxson signed-rank test depicts LSTM-FCN with dimension shuffle to statistically outperform GRU-FCN, RNN-FCN, Dense-FCN. Surprisingly, the model to perform most similar to LSTM-FCN with dimension shuffle is Dense-FCN. LSTM-FCN outperforms Dense-FCN in 231 experiments, ties in 35 experiments and performs worse in 118 experiments. 

An interesting observation is that GRU-FCN does not statistically outperform Dense-FCN. Based on our 384 experiments, GRU-FCN outpeforms Dense-FCN in 160 experiments, ties in 49 experiments, while performing worse in 175 experiments. As a disclaimer, we performed each of these experiments only once, therefore there may be some deviation when run multiple times due to the inherent variance of training using random initialization. However, due to the sample size of 384, we believe the variance will not be significant to result in a different conclusion.

\section{Conclusion \& Future Work}
\label{conclusion}
In this paper, we provide a better understanding of LSTM-FCN, ALSTM-FCN and their sub-modules through a series of ablation tests (3627 experiments). We show that \textit{z-normalizing} the whole dataset yields to results different than \textit{z-normalizing} each sample. For the model LSTM-FCN and ALSTM-FCN, we recommend \textit{z-normalizing} the whole dataset only in situations when it is known that the training set is a good representation of the global population. Moreover, our ablation tests show that the LSTM/ALSTM block and the FCN block yields to a better performing model when applied in a conjoined manner. Further, the performance of LSTM-FCN is enhanced only when dimension shuffle is applied before the LSTM block. Finally, in this paper, we substitute the LSTM block with either a GRU block, a RNN block or a Dense block to observe the effect of such a substitution. Our results indicate LSTM-FCN to outperform GRU-FCN, RNN-FCN and Dense-FCN.

An exciting area for future work is to investigate why LSTM-FCN and ALSTM-FCN underperform in a few UCR datasets and to ascertain whether the models can be made more robust to the various types of time series data. Furthermore, integrating the models in both low-power systems and wearables for on-device classification is of great interest. Finally, further inroads can be made in streaming time series classification by the utilization of these models. \hl{In the future, researchers that want to implement deep learning models for time series classification need to focus on generalization of the model on unseen sequences, and reduce over-fitting as the UCR repository contain small real world data sets.}


%


\section*{Acknowledgment}

The authors would like to thank all the researchers that helped create and clean the data available in the updated UCR Time Series Classification Archive. We would also like to show our gratitude to the administrators of the UCR Time Series Classification Archive, Dau et al. Sustained research in this domain would be much more challenging without their efforts.
\par Further, the authors would like to acknowledge the Research Open Access Publishing (ROAAP) Fund of the University of Illinois at Chicago for financial support towards the open access publishing fee for this article.  

\ifCLASSOPTIONcaptionsoff
  \newpage
\fi



\bibliographystyle{IEEEtran}

\bibliography{biblio}{}

\begin{thebibliography}{10}
\providecommand{\url}[1]{#1}
\csname url@samestyle\endcsname
\providecommand{\newblock}{\relax}
\providecommand{\bibinfo}[2]{#2}
\providecommand{\BIBentrySTDinterwordspacing}{\spaceskip=0pt\relax}
\providecommand{\BIBentryALTinterwordstretchfactor}{4}
\providecommand{\BIBentryALTinterwordspacing}{\spaceskip=\fontdimen2\font plus
\BIBentryALTinterwordstretchfactor\fontdimen3\font minus
  \fontdimen4\font\relax}
\providecommand{\BIBforeignlanguage}[2]{{%
\expandafter\ifx\csname l@#1\endcsname\relax
\typeout{** WARNING: IEEEtran.bst: No hyphenation pattern has been}%
\typeout{** loaded for the language `#1'. Using the pattern for}%
\typeout{** the default language instead.}%
\else
\language=\csname l@#1\endcsname
\fi
#2}}
\providecommand{\BIBdecl}{\relax}
\BIBdecl

\bibitem{dau2018ucr}
H.~A. Dau, A.~Bagnall, K.~Kamgar, C.-C.~M. Yeh, Y.~Zhu, S.~Gharghabi, C.~A.
  Ratanamahatana, and E.~Keogh, ``The ucr time series archive,'' \emph{arXiv
  preprint arXiv:1810.07758}, 2018.

\bibitem{sharabiani2017efficient}
A.~Sharabiani, H.~Darabi, A.~Rezaei, S.~Harford, H.~Johnson, and F.~Karim,
  ``Efficient classification of long time series by 3-d dynamic time warping,''
  \emph{IEEE Transactions on Systems, Man, and Cybernetics: Systems}, 2017.

\bibitem{Sharabiani2014_bayesian}
A.~Sharabiani, F.~Karim, A.~Sharabiani, M.~Atanasov, and H.~Darabi, ``An
  enhanced bayesian network model for prediction of students' academic
  performance in engineering programs,'' in \emph{2014 IEEE Global Engineering
  Education Conference (EDUCON)}, April 2014, pp. 832--837.

\bibitem{karim2018multivariate}
F.~Karim, S.~Majumdar, H.~Darabi, and S.~Harford, ``Multivariate lstm-fcns for
  time series classification,'' \emph{arXiv preprint arXiv:1801.04503}, 2018.

\bibitem{wei2006semi}
L.~Wei and E.~Keogh, ``Semi-supervised time series classification,'' in
  \emph{Proceedings of the 12th ACM SIGKDD international conference on
  Knowledge discovery and data mining}.\hskip 1em plus 0.5em minus 0.4em\relax
  ACM, 2006, pp. 748--753.

\bibitem{taylor2009wind}
J.~W. Taylor, P.~E. McSharry, and R.~Buizza, ``Wind power density forecasting
  using ensemble predictions and time series models,'' \emph{IEEE Transactions
  on Energy Conversion}, vol.~24, no.~3, pp. 775--782, 2009.

\bibitem{tsay2005analysis}
R.~S. Tsay, \emph{Analysis of financial time series}.\hskip 1em plus 0.5em
  minus 0.4em\relax John Wiley \& Sons, 2005, vol. 543.

\bibitem{sternickel2002automatic}
K.~Sternickel, ``Automatic pattern recognition in ecg time series,''
  \emph{Computer methods and programs in biomedicine}, vol.~68, no.~2, pp.
  109--115, 2002.

\bibitem{theiler1992testing}
J.~Theiler, S.~Eubank, A.~Longtin, B.~Galdrikian, and J.~D. Farmer, ``Testing
  for nonlinearity in time series: the method of surrogate data,''
  \emph{Physica D: Nonlinear Phenomena}, vol.~58, no. 1-4, pp. 77--94, 1992.

\bibitem{maharaj2019time}
E.~A. Maharaj, P.~D'Urso, and J.~Caiado, \emph{Time Series Clustering and
  Classification}.\hskip 1em plus 0.5em minus 0.4em\relax CRC Press, 2019.

\bibitem{Lea_2016}
C.~Lea, R.~Vidal, A.~Reiter, and G.~D. Hager, ``{Temporal Convolutional
  Networks: A Unified Approach to Action Segmentation},'' in \emph{Lecture
  Notes in Computer Science}.\hskip 1em plus 0.5em minus 0.4em\relax Springer
  International Publishing, 2016, pp. 47--54.

\bibitem{xi2006fast}
X.~Xi, E.~Keogh, C.~Shelton, L.~Wei, and C.~A. Ratanamahatana, ``Fast time
  series classification using numerosity reduction,'' in \emph{Proceedings of
  the 23rd International Conference on Machine Learning}.\hskip 1em plus 0.5em
  minus 0.4em\relax ACM, 2006, pp. 1033--1040.

\bibitem{jain2018asymmetric}
B.~J. Jain and D.~Schultz, ``Asymmetric learning vector quantization for
  efficient nearest neighbor classification in dynamic time warping spaces,''
  \emph{Pattern Recognition}, vol.~76, pp. 349--366, 2018.

\bibitem{agrawal1993efficient}
R.~Agrawal, C.~Faloutsos, and A.~Swami, ``Efficient similarity search in
  sequence databases,'' in \emph{International conference on foundations of
  data organization and algorithms}.\hskip 1em plus 0.5em minus 0.4em\relax
  Springer, 1993, pp. 69--84.

\bibitem{keogh2003need}
E.~Keogh and S.~Kasetty, ``On the need for time series data mining benchmarks:
  a survey and empirical demonstration,'' \emph{Data Mining and knowledge
  discovery}, vol.~7, no.~4, pp. 349--371, 2003.

\bibitem{UCRArchive}
Y.~Chen, E.~Keogh, B.~Hu, N.~Begum, A.~Bagnall, A.~Mueen, and G.~Batista,
  ``{The UCR Time Series Classification Archive},'' July 2015,
  \url{www.cs.ucr.edu/~eamonn/time_series_data/}.

\bibitem{UCRArchive2018}
H.~A. Dau, E.~Keogh, K.~Kamgar, C.-C.~M. Yeh, Y.~Zhu, S.~Gharghabi, C.~A.
  Ratanamahatana, Yanping, B.~Hu, N.~Begum, A.~Bagnall, A.~Mueen, and
  G.~Batista, ``The ucr time series classification archive,'' October 2018,
  \url{https://www.cs.ucr.edu/~eamonn/time_series_data_2018/}.

\bibitem{Lin_2007}
J.~Lin, E.~Keogh, L.~Wei, and S.~Lonardi, ``{Experiencing {SAX}: A Novel
  Symbolic Representation of Time Series},'' \emph{Data Mining and Knowledge
  Discovery}, vol.~15, no.~2, pp. 107--144, apr 2007.

\bibitem{Baydogan_2013}
M.~G. Baydogan, G.~Runger, and E.~Tuv, ``{A Bag-of-Features Framework to
  Classify Time Series},'' \emph{{IEEE} Transactions on Pattern Analysis and
  Machine Intelligence}, vol.~35, no.~11, pp. 2796--2802, nov 2013.

\bibitem{Sch_fer_2014}
P.~Sch{\"a}fer, ``{The {BOSS} is Concerned with Time Series Classification in
  the Presence of Noise},'' \emph{Data Mining and Knowledge Discovery},
  vol.~29, no.~6, pp. 1505--1530, sep 2014.

\bibitem{schafer2016scalable}
P.~Sch{\"a}fer, ``{Scalable Time Series Classification},'' \emph{Data Mining
  and Knowledge Discovery}, vol.~30, no.~5, pp. 1273--1298, 2016.

\bibitem{Schafer_2017}
P.~Sch{\"a}fer and U.~Leser, ``{Fast and Accurate Time Series Classification
  with WEASEL},'' \emph{arXiv preprint arXiv:1701.07681}, 2017.

\bibitem{Lines_2014}
J.~Lines and A.~Bagnall, ``{Time Series Classification with Ensembles of
  Elastic Distance Measures},'' \emph{Data Mining and Knowledge Discovery},
  vol.~29, no.~3, pp. 565--592, jun 2014.

\bibitem{bagnall2015time}
A.~Bagnall, J.~Lines, J.~Hills, and A.~Bostrom, ``{Time-Series Classification
  with COTE: The Collective of Transformation-Based Ensembles},'' \emph{IEEE
  Transactions on Knowledge and Data Engineering}, vol.~27, no.~9, pp.
  2522--2535, 2015.

\bibitem{cui2016multi}
Z.~Cui, W.~Chen, and Y.~Chen, ``{Multi-Scale Convolutional Neural Networks for
  Time Series Classification},'' \emph{arXiv preprint arXiv:1603.06995}, 2016.

\bibitem{wang2017time}
Z.~Wang, W.~Yan, and T.~Oates, ``{Time Series Classification from Scratch with
  Deep Neural Networks: A Strong Baseline},'' in \emph{Neural Networks (IJCNN),
  2017 International Joint Conference on}.\hskip 1em plus 0.5em minus
  0.4em\relax IEEE, 2017, pp. 1578--1585.

\bibitem{karim2018lstm}
F.~Karim, S.~Majumdar, H.~Darabi, and S.~Chen, ``Lstm fully convolutional
  networks for time series classification,'' \emph{IEEE Access}, vol.~6, pp.
  1662--1669, 2018.

\bibitem{kim2018resource}
Y.~Kim, J.~Sa, Y.~Chung, D.~Park, and S.~Lee, ``Resource-efficient pet dog
  sound events classification using lstm-fcn based on time-series data,''
  \emph{Sensors}, vol.~18, no.~11, p. 4019, 2018.

\bibitem{ioffe2015batch}
S.~Ioffe and C.~Szegedy, ``{Batch Normalization: Accelerating Deep Network
  Training by Reducing Internal Covariate Shift},'' in \emph{International
  Conference on Machine Learning}, 2015, pp. 448--456.

\bibitem{Trottier2016}
\BIBentryALTinterwordspacing
L.~Trottier, P.~Gigu{\`{e}}re, and B.~Chaib-draa, ``{Parametric Exponential
  Linear Unit for Deep Convolutional Neural Networks},'' \emph{arXiv}, pp.
  1--16, may 2016. [Online]. Available: \url{http://arxiv.org/abs/1605.09332}
\BIBentrySTDinterwordspacing

\bibitem{pascanu2013construct}
R.~Pascanu, C.~Gulcehre, K.~Cho, and Y.~Bengio, ``{How to Construct Deep
  Recurrent Neural Networks},'' \emph{arXiv preprint arXiv:1312.6026}, 2013.

\bibitem{hochreiter1997long}
S.~Hochreiter and J.~Schmidhuber, ``{Long Short-Term Memory},'' \emph{Neural
  computation}, vol.~9, no.~8, pp. 1735--1780, 1997.

\bibitem{graves2012supervised}
A.~Graves \emph{et~al.}, \emph{{Supervised Sequence Labelling with Recurrent
  Neural Networks}}.\hskip 1em plus 0.5em minus 0.4em\relax Springer, 2012,
  vol. 385.

\bibitem{cho2014gru}
K.~Cho, B.~Van~Merri{\"e}nboer, C.~Gulcehre, D.~Bahdanau, F.~Bougares,
  H.~Schwenk, and Y.~Bengio, ``Learning phrase representations using rnn
  encoder-decoder for statistical machine translation,'' \emph{arXiv preprint
  arXiv:1406.1078}, 2014.

\bibitem{king2001logistic}
G.~King and L.~Zeng, ``{Logistic Regression in Rare Events Data},''
  \emph{Political analysis}, vol.~9, no.~2, pp. 137--163, 2001.

\bibitem{chollet2015keras}
F.~Chollet \emph{et~al.}, ``Keras,'' \url{https://github.com/fchollet/keras},
  2015.

\bibitem{tensorflow2015-whitepaper}
\BIBentryALTinterwordspacing
M.~Abadi, A.~Agarwal, P.~Barham, E.~Brevdo, Z.~Chen, C.~Citro, G.~S. Corrado,
  A.~Davis, J.~Dean, M.~Devin, S.~Ghemawat, I.~Goodfellow, A.~Harp, G.~Irving,
  M.~Isard, Y.~Jia, R.~Jozefowicz, L.~Kaiser, M.~Kudlur, J.~Levenberg,
  D.~Man\'{e}, R.~Monga, S.~Moore, D.~Murray, C.~Olah, M.~Schuster, J.~Shlens,
  B.~Steiner, I.~Sutskever, K.~Talwar, P.~Tucker, V.~Vanhoucke, V.~Vasudevan,
  F.~Vi\'{e}gas, O.~Vinyals, P.~Warden, M.~Wattenberg, M.~Wicke, Y.~Yu, and
  X.~Zheng, ``{TensorFlow}: Large-scale machine learning on heterogeneous
  systems,'' 2015, software available from tensorflow.org. [Online]. Available:
  \url{https://www.tensorflow.org/}
\BIBentrySTDinterwordspacing

\bibitem{kingma2014adam}
D.~Kingma and J.~Ba, ``{Adam: A Method for Stochastic Optimization},''
  \emph{arXiv preprint arXiv:1412.6980}, 2014.

\bibitem{Krizhevsky2012Imagenet}
\BIBentryALTinterwordspacing
A.~Krizhevsky, I.~Sutskever, and G.~E. Hinton, ``Imagenet classification with
  deep convolutional neural networks,'' in \emph{Advances in Neural Information
  Processing Systems 25}, F.~Pereira, C.~J.~C. Burges, L.~Bottou, and K.~Q.
  Weinberger, Eds.\hskip 1em plus 0.5em minus 0.4em\relax Curran Associates,
  Inc., 2012, pp. 1097--1105. [Online]. Available:
  \url{http://papers.nips.cc/paper/4824-imagenet-classification-with-deep-convolutional-neural-networks.pdf}
\BIBentrySTDinterwordspacing

\bibitem{he2015delving}
K.~He, X.~Zhang, S.~Ren, and J.~Sun, ``{Delving Deep into Rectifiers:
  Surpassing Human-Level Performance on Imagenet Classification},'' in
  \emph{Proceedings of the IEEE international conference on computer vision},
  2015, pp. 1026--1034.

\bibitem{Szegedy_2016_CVPR}
C.~Szegedy, V.~Vanhoucke, S.~Ioffe, J.~Shlens, and Z.~Wojna, ``Rethinking the
  inception architecture for computer vision,'' in \emph{The IEEE Conference on
  Computer Vision and Pattern Recognition (CVPR)}, June 2016.

\bibitem{szegedy2017inception}
C.~Szegedy, S.~Ioffe, V.~Vanhoucke, and A.~A. Alemi, ``Inception-v4,
  inception-resnet and the impact of residual connections on learning.'' in
  \emph{AAAI}, vol.~4, 2017, p.~12.

\bibitem{hu2018senet}
J.~Hu, L.~Shen, and G.~Sun, ``Squeeze-and-excitation networks,'' in \emph{IEEE
  Conference on Computer Vision and Pattern Recognition}, 2018.

\bibitem{wilcoxon1964some}
F.~Wilcoxon and R.~A. Wilcox, \emph{Some rapid approximate statistical
  procedures}.\hskip 1em plus 0.5em minus 0.4em\relax Lederle Laboratories,
  1964.

\bibitem{abdi2007bonferroni}
H.~Abdi, ``Bonferroni and {\v{s}}id{\'a}k corrections for multiple
  comparisons,'' \emph{Encyclopedia of measurement and statistics}, vol.~3, pp.
  103--107, 2007.

\bibitem{schafer2017fast}
P.~Sch{\"a}fer and U.~Leser, ``{Fast and Accurate Time Series Classification
  with WEASEL},'' \emph{arXiv preprint arXiv:1701.07681}, 2017.

\bibitem{elsayed2018deep}
N.~Elsayed, A.~S. Maida, and M.~Bayoumi, ``Deep gated recurrent and
  convolutional network hybrid model for univariate time series
  classification,'' \emph{arXiv preprint arXiv:1812.07683}, 2018.

\end{thebibliography}
%



%





\end{document}